\def\cvprPaperID{6444}
\def\confName{CVPR}
\def\confYear{2023}

\def\paperTitle{Explicit Visual Prompting for Low-Level Structure Segmentations}

\def\authorBlock{
    Weihuang Liu\textsuperscript{~\rm 1} \qquad
    Xi Shen\textsuperscript{~\rm 2} \qquad
    Chi-Man Pun\textsuperscript{~\rm 1}\thanks{Corresponding Author} \qquad
    Xiaodong Cun\textsuperscript{~\rm 2}\footnotemark[1] \\
    \textsuperscript{\rm 1}~University of Macau \qquad 
    \textsuperscript{\rm 2}~Tencent AI Lab \\ \\
}

\newif\ifreview 
\newif\ifarxiv \newcommand{\arxiv}{\arxivtrue}
\newif\ifcamera 
\newif\ifrebuttal 

\arxiv 

\pdfoutput=1
\documentclass[10pt,twocolumn,letterpaper]{article}
\ifreview \usepackage[review]{cvpr} \fi
\ifarxiv \usepackage[pagenumbers]{cvpr} \fi
\ifrebuttal \usepackage[rebuttal]{cvpr} \fi
\ifcamera \usepackage{cvpr} \fi

\usepackage{graphicx}
\usepackage{amsmath}
\usepackage{amssymb}
\usepackage{booktabs}

\usepackage{times}
\usepackage{microtype}
\usepackage{epsfig}
\usepackage[table,xcdraw]{xcolor}
\usepackage{float}
\usepackage{placeins}
\usepackage{color, colortbl}
\usepackage{stfloats}
\usepackage{enumitem}
\usepackage{tabularx}
\usepackage{xstring}
\usepackage{multirow}
\usepackage{xspace}
\usepackage{url}
\usepackage{subcaption}
\usepackage{xcolor}

\ifcamera \usepackage[accsupp]{axessibility} \fi

\definecolor{asparagus}{rgb}{0.04, 0.85, 0.32}

\ifarxiv  \fi

\newcommand{\R}[1]{{%
    \textbf{%
        \ifstrequal{#1}{1}{\textcolor{red}{R#1}}{%
        \ifstrequal{#1}{2}{\textcolor{blue}{R#1}}{%
        \ifstrequal{#1}{3}{\textcolor{magenta}{R#1}}{%
        \ifstrequal{#1}{4}{\textcolor{teal}{R#1}}{%
                           \textcolor{cyan}{R#1}%
        }}}}%
    }%
}}

\usepackage{xr-hyper}

\makeatletter
\newcommand*{\addFileDependency}[1]{
  \typeout{(#1)}
  \@addtofilelist{#1}
  \IfFileExists{#1}{}{\typeout{No file #1.}}
}

\makeatother

\usepackage[pagebackref,breaklinks,colorlinks]{hyperref}
\usepackage[capitalize]{cleveref}
\crefname{section}{Sec.}{Secs.}
\crefname{table}{Table}{Tables}
\crefname{figure}{Fig.}{Figs.}

\hypersetup{
    colorlinks = true,
    linkbordercolor = {white},
    citecolor=blue
}

\begin{document}
\title{\paperTitle}
\author{\authorBlock}
\maketitle

\begin{abstract}
We consider the generic problem of detecting low-level structures in images, which includes segmenting the manipulated parts, identifying out-of-focus pixels, separating shadow regions, and detecting concealed objects.
Whereas each such topic has been typically addressed with a domain-specific solution, we show that a unified approach performs well across all of them. 
We take inspiration from the widely-used pre-training and then prompt tuning protocols in NLP and propose a new visual prompting model, named Explicit Visual Prompting~(EVP). 
Different from the previous visual prompting which is typically a dataset-level implicit embedding, our key insight is to enforce the tunable parameters focusing on the explicit visual content from each individual image, \ie, the features from frozen patch embeddings and the input's high-frequency components.
The proposed EVP significantly outperforms other parameter-efficient tuning protocols under the same amount of tunable parameters~(5.7$\%$ extra trainable parameters of each task). EVP also achieves state-of-the-art performances on diverse low-level structure segmentation tasks compared to task-specific solutions. 
Our code is available at:~\url{https://github.com/NiFangBaAGe/Explicit-Visual-Prompt}.

\end{abstract}

\section{Introduction}
\label{sec:intro}
\begin{figure}[tp]
    \centering
    \includegraphics[width=\linewidth]{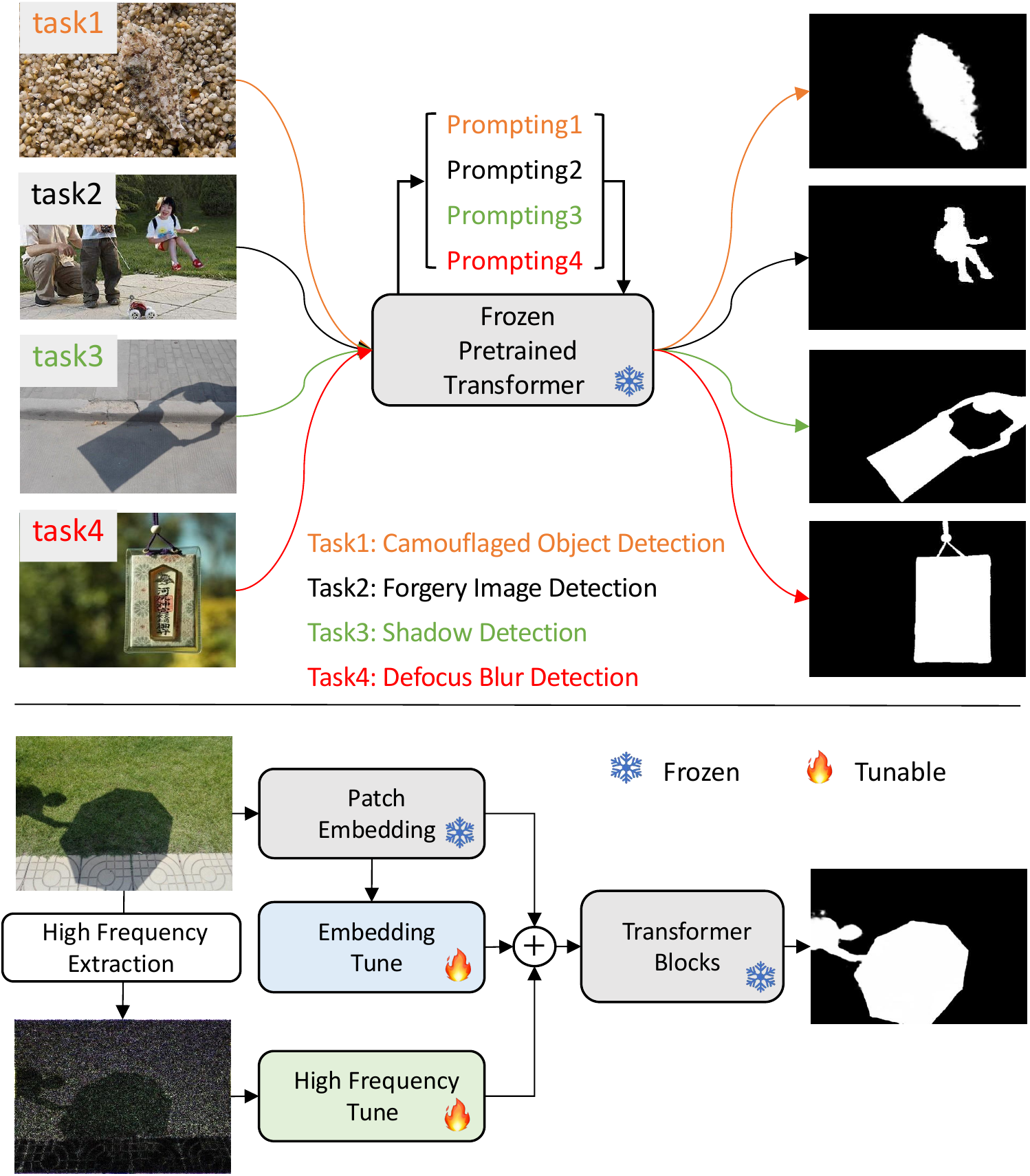}
    \caption{We propose a unified method for four low-level structure segmentation tasks: camouflaged object, forgery, shadow and defocus blur detection~(Top). Our approach relies on a pre-trained frozen transformer backbone that leverages explicit extracted features, \eg, the frozen embedded features and high-frequency components, to prompt knowledge. } 
    \label{fig:teaser}
\end{figure}

Advances in image editing and manipulation algorithms have made it easy to create photo-realistic but fake pictures~\cite{huh2018fighting,rombach2022high,kawar2022imagic}. Detecting such manipulated regions becomes an important problem due to its potential negative impact related to surveillance and crime~\cite{huh2018fighting}. Low-level structures are known to be beneficial to tampered region detection, \ie, resizing and copy-pasting will destroy the JPEG compression levels between the temper region and the host image~\cite{luo2010jpeg,huang2010detecting,popescu2005exposing}, the noise level of the tempered region and the background is also different~\cite{zhou2018learning, wu2019mantra}.
Interesting, to segment the blurred pixels~\cite{shi2014discriminative}, shadowed regions~\cite{panagopoulos2011illumination}, and concealed objects~\cite{fan2020camouflaged}, low-level clues also play important roles. 
These detection tasks are shown to be beneficial to numerous computer vision tasks, including auto-refocus~\cite{bae2007defocus}, image retargeting~\cite{karaali2016image}, object tracking~\cite{mikic2000moving}, \etc.

Although all these tasks belong to low-level structure segmentation, they are typically addressed by domain-specific solutions with carefully designed network architectures~\cite{zhou2018learning,cun2020defocus,zhu2021mitigating}. Moreover, the lack of large-scale datasets is often considered a major factor, which limits the performances~\cite{huh2018fighting}.

In this work, we propose a solution to address the four tasks in a unified fashion. We take inspiration from recent advances of \textit{prompting}~\cite{vpt,chen2022adaptformer,bar2022visual}, which is a concept that initially emerged in natural language processing (NLP)~\cite{brown2020language}. The basic idea is to efficiently adapt a frozen large foundation model to many downstream tasks with the minimum extra trainable parameters. As the foundation model has already been trained on a large-scale dataset, prompting often leads to better model generalization on the downstream tasks~\cite{brown2020language}, especially in the case of the limited annotated data. Prompting also significantly saves the storage of models since it only needs to save a shared basic model and task-aware promptings.

Our main insight is to tune the task-specific knowledge \textit{only} from the features of each individual image itself because the pre-trained base model contains sufficient knowledge for semantic understanding. This is also inspired by the effectiveness of hand-crafted image features, such as SIFT~\cite{huang2008detection}, JPEG noise~\cite{luo2010jpeg}, resampling artifacts~\cite{popescu2005exposing} in these tasks~\cite{liu2011detection,luo2010jpeg,huang2008detection,huang2010detecting,popescu2005exposing,zhou2018learning}.

Based on this observation, we propose \textit{explicit visual prompting~(EVP)}, where the tuning performance can be hugely improved via the re-modulation of image features. Specifically, we consider two kinds of features for our task. 
The first is the features from the frozen patch embedding, which is critical since we need to shift the distribution of the original model.
Another is high-frequency components of the input image since the pre-trained visual recognition model is learned to be invariant to these features via data augmentation. 
As shown in Figure~\ref{fig:teaser}, we take a model pre-trained on a large-scale dataset and freeze its parameters. Then, to adapt to each task, we tune the embedded features and learn an extra embedding for high-frequency components of each individual image.

In terms of experiments, we validate our approach on nine datasets of four tasks: forgery detection, shadow detection, defocus blur detection as well as camouflaged object detection. Our simple and unified network achieves very competitive performance with the whole model fine-tuning and outperforms task-specific solutions without modification.

In summary, our main contributions are as follows:
\begin{itemize}
    \item We design a unified approach that produces state-of-the-art performances for a number of tasks, including forgery detection, defocus blur detection, shadow detection, and camouflaged object detection. 
    
    \item We propose explicit visual prompting (EVP), which takes the features from the frozen patch embedding and the input's high-frequency components as prompting. It is demonstrated to be effective across different tasks and outperforms other parameter-efficient tuning methods. 
    
    \item Our method greatly simplifies the low-level structure segmentation models as well as achieves comparable performance with well-designed SOTA methods.
    
\end{itemize}
\section{Related Work} 

\label{sec:related}

\paragraph{Visual Prompting Tuning.} Prompting is initially proposed in NLP~\cite{brown2020language,liu2021pre}.~\cite{brown2020language} demonstrates strong generalization to downstream transfer learning
tasks even in the few-shot or zero-shot settings with manually chosen
prompts in GPT-3. Recently, prompting~\cite{sandler2022fine,vpt} has been adapted to vision tasks.~\cite{sandler2022fine} proposes memory tokens which is a set of learnable embedding vectors for each transformer layer. VPT~\cite{vpt} proposes similar ideas and investigates the generality and feasibility of visual prompting
via extensive experiments spanning multiple kinds of recognition tasks across multiple domains and backbone architectures. Unlike VPT, whose main focus is on recognition tasks, our work aims at exploring optimal visual content for low-level structure segmentation. 

\paragraph{Forgery Detection.} The goal of forgery detection is to detect pixels that are manually manipulated, such as pixels that are removed, replaced, or edited. Early approaches~\cite{mahdian2009using,lyu2014exposing,fridrich2012rich, cun2018image} detect region splicing through inconsistencies in local noise levels, based on the fact that images of different origins might contain different noise characteristics introduced by the sensors or post-processing steps. Other clues are found to be helpful, such as SIFT~\cite{huang2008detection}, JPEG compression artifacts~\cite{luo2010jpeg} and re-sampling artifacts~\cite{feng2012normalized,popescu2005exposing}. Recently, approaches have moved towards end-to-end deep learning methods for solving specific forensics tasks using labeled training data~\cite{islam2020doa,wu2017deep,zhong2019end,salloum2018image,hu2020span}. Salloum 
\etal~\cite{salloum2018image} learn to detect splicing by training a fully convolutional network on labeled training data.~\cite{wu2017deep,zhong2019end,wu2019mantra,hu2020span,psccnet} propose improved architectures. Islam \etal~\cite{islam2020doa} incorporate Generative Adversarial Network (GAN) to detect copy-move forgeries. Huh \etal~\cite{huh2018fighting} propose to take photographic metadata as a free and plentiful supervisory signal for learning self-consistency and apply the trained model to detect splices. Recently, TransForensic~\cite{hao2021transforensics} leverages vision transformers~\cite{dosovitskiy2020image} to tackle the problem. High-frequency components still served as useful prior in this field. RGB-N~\cite{zhou2018learning} designs an additional noise stream. ObjectFormer~\cite{wang2022objectformer} extracts high-frequency features as complementary signals to visual content. But unlike ObjectFormer, our main focus is to leverage high-frequency components as a prompting design to efficiently and effectively adapt to different low-level segmentation tasks.

\paragraph{Defocus Blur Detection.} Given an image, defocus blur detection aims at separating in-focus and out-of-focus regions, which could be potentially useful for auto-refocus~\cite{bae2007defocus}, salient object detection~\cite{jiang2013salient} and image retargeting~\cite{karaali2016image}. Traditional approaches mainly focus on designing hand-crafted features based on gradient~\cite{shi2014discriminative,yi2016lbp,golestaneh2017hifst} or edge~\cite{karaali2017edge,shi2015just}. In the deep era, most methods delve into CNN architectures~\cite{park2017unified,zhao2019btbnet,tang2019defusionnet,zhao2019cenet}.~\cite{park2017unified} proposes the first CNN-based method using both hand-crafted and deep features. BTBNet~\cite{zhao2019btbnet} develops a fully convolutional network to integrate low-level clues and high-level semantic information. DeFusionNet~\cite{tang2019defusionnet} recurrently fuses and refines multi-scale deep features for defocus blur detection. CENet~\cite{zhao2019cenet} learns multiple smaller defocus blur detectors and ensembles them to enhance diversity.~\cite{cun2020defocus} further employs the depth information as additional supervision and proposes a joint learning framework inspired by knowledge distillation.~\cite{zhao2021defocus} explores deep ensemble networks for defocus blur detection.~\cite{zhao2021self} proposes to learn generator to generate mask in an adversarial manner.

\paragraph{Shadow Detection.} Shadows occur frequently in natural scenes, and have hints for scene geometry~\cite{okabe2009attached}, light conditions~\cite{okabe2009attached} and camera location ~\cite{junejo2008estimating} and lead to challenging cases in many vision tasks including image segmentation~\cite{ecins2014shadow} and object tracking~\cite{cucchiara2003detecting,nadimi2004physical}.
Early attempts explore illumination ~\cite{finlayson2005removal,finlayson2009entropy} and hand-crafted features~\cite{huang2011characterizes,lalonde2010detecting,zhu2010learning}. In the deep era, some methods mainly focus on the design of CNN architectures~\cite{zhu2018bidirectional,cun2020towards} or involving the attention modules~(\emph{e.g.}, the direction-aware attention~\cite{hu2018direction}, distraction-aware module~\cite{zheng2019distraction}). Recent works~\cite{le2018a+,zhu2021mitigating} utilize the lighting as additional prior, for example, ADNet~\cite{le2018a+} generates the adversarial training samples for better detection and FDRNet~\cite{zhu2021mitigating} arguments the training samples by additionally adjusted brightness. MTMT~\cite{mtmt} leverages the mean teacher model to explore unlabeled data for semi-supervised shadow detection.

\paragraph{Camouflaged Object Detection.} 
Detecting camouflaged objects is a challenging task as foreground objects are often with visual similar patterns to the background. Early works distinguish the foreground and background through low-level clues such as texture~\cite{sengottuvelan2008performance,feng2013camouflage}, brightness~\cite{pike2018quantifying}, and color~\cite{hou2011detection}. Recently, deep learning-based methods~\cite{fan2020camouflaged,mei2021camouflaged,li2021uncertainty,lv2021simultaneously,jiaying2022frequency} show their strong ability in detecting complex camouflage objects. Le \etal~\cite{le2019anabranch} propose the first end-to-end network for camouflaged object detection, which is composed of a classification branch and a segmentation branch. Fan \etal~\cite{fan2020camouflaged} develops a search-identification network and the largest camouflaged object detection dataset. PFNet~\cite{mei2021camouflaged} is a bio-inspired framework that mimics the process of positioning and identification in predation. FBNet~\cite{jiaying2022frequency} suggests disentangling frequency modeling and enhancing the important frequency component.

\section{Method}
\label{sec:method}
In this section, we propose Explicit Visual Prompting~(EVP) for adapting recent Vision Transformers~(SegFormer~\cite{xie2021segformer} as the example) pre-trained on ImageNet~\cite{deng2009imagenet} to low-level structure segmentations. EVP keeps the backbone frozen and only contains a small number of tunable parameters to learn task-specific knowledge from the features of frozen image embeddings and high-frequency components. Below, we first present SegFormer\cite{xie2021segformer} and the extraction of high-frequency components in Section~\ref{Preliminary}, then the architecture design in Section~\ref{sec:explicit_visual_prompting}.

\subsection{Preliminaries}
\label{Preliminary}

\paragraph{SegFormer~\cite{xie2021segformer}.} SegFormer is a hierarchical transformer-based structure with a much simpler decoder for semantic segmentation. Similar to traditional CNN backbone~\cite{resnet}, SegFormer captures multi-stale features via several stages. Differently, each stage is built via the feature embedding layers\footnote{SegFormer has a different definition of patch embedding in ViT~\cite{dosovitskiy2020image}. It uses the overlapped patch embedding to extract the denser features and will merge the embedding to a smaller spatial size at the beginning of each stage.} and vision transformer blocks~\cite{vaswani2017attention, dosovitskiy2020image}. As for the decoder, it leverages the multi-scale features from the encoder and MLP layers for decoding to the specific classes. Notice that, the proposed prompt strategy is not limited to SegFormer and can be easily adapted to other network structures, \eg, ViT~\cite{dosovitskiy2020image} and Swin~\cite{liu2021swin}.


\begin{figure}[tp]
    \centering
    \includegraphics[width=\linewidth]{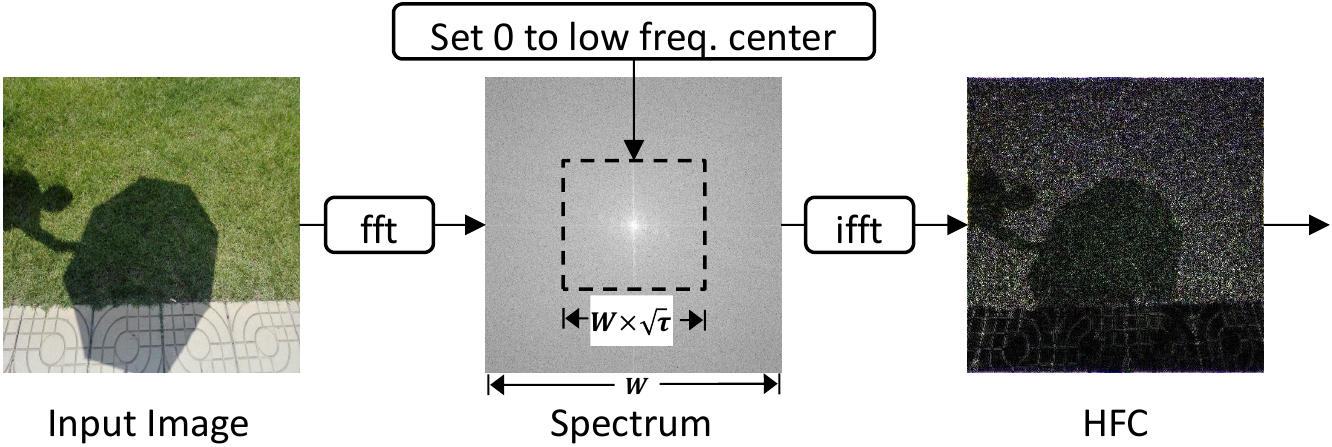}
    \vspace{-1em}
    \caption{The procedure to generate high-frequency components.}
    \label{fig:hfc}
\end{figure}

\paragraph{High-frequency Components (HFC).} As shown in Figure~\ref{fig:hfc}, for an image $I$ of dimension $H \times W$, we can decompose it into low-frequency components $I_l$ (LFC) and high-frequency components $I_h$ (HFC), \textit{i.e.} $I = \{I_l, I_h\}$. Denoting $\mathtt{fft}$ and $\mathtt{ifft}$ as the Fast Fourier Transform and its inverse respectively, we use $z$ to represent the frequency component of $I$. Therefore we have $z = \mathtt{fft}(I)$ and $I = \mathtt{ifft}(z)$. We shift low frequency coefficients to the center $(\frac{H}{2}, \frac{W}{2})$. To obtain HFC, a binary mask $\mathbf{M}_h \in \{0, 1\}^{H \times W}$ is generated and applied on $z$ depending on a mask ratio $\tau$: 

\begin{equation}
\mathbf{M}^{i,j}_h(\tau) = \begin{cases} 
	0, & \frac{4|(i - \frac{H}{2})(j - \frac{W}{2})|}{HW} \le \tau\\ 
	1, & \text{otherwise} 
\end{cases}
\end{equation}

$\tau$ indicates the surface ratio of the masked regions. HFC can be computed:
\begin{align}
	I_{hfc} =  \mathtt{ifft}(z \mathbf{M}_h(\tau))
\end{align}

Similarly, a binary mask $\mathbf{M}_l \in \{0, 1\}^{H \times W}$ can be properly defined to compute LFC: 
\begin{equation}
	\mathbf{M}^{i,j}_l(\tau) = \begin{cases} 
		0, & \frac{HW - 4|(i - \frac{H}{2})(j - \frac{W}{2})|}{HW} \le \tau\\ 
		1, & \text{otherwise} 
	\end{cases}
\end{equation}
and LFC can be calculated as:
\begin{align}
	I_{lfc} =  \mathtt{ifft}(z \mathbf{M}_l(\tau))
\end{align}

Note that for RGB images, we compute the above process on every channel of pixels independently.

\begin{figure}[tp]
    \centering
    \includegraphics[width=\linewidth]{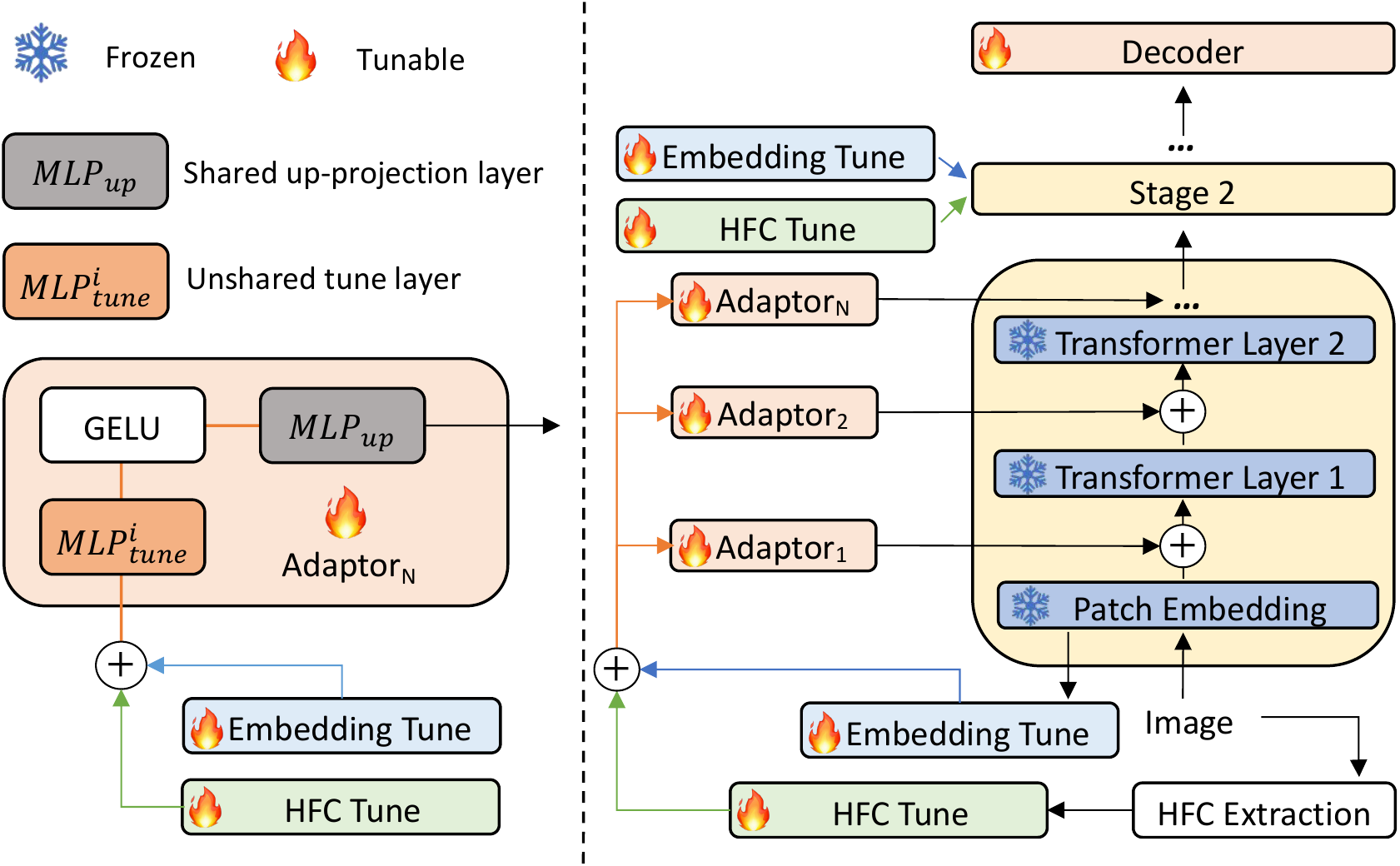}
    \vspace{-1em}
    \caption{The architecture of the proposed explicit visual prompting. We use the \textit{Embedding Tune} and the \textit{HFC Tune} to tune the extracted features. The Adaptor is designed to merge these features. }
    \label{fig:arch}
\end{figure}

\subsection{Explicit Visual Prompting}
\label{sec:explicit_visual_prompting}

In this section, we present the proposed Explicit Visual Prompting (EVP). Our key insight is to learn explicit prompts from image embeddings and high-frequency components.
We learn the former to shift the distribution from the pre-train dataset to the target dataset. 
And the main motivation to learn the latter is that the pre-trained model is learned to be invariant to these features through data augmentation.
Note that this is different from VPT~\cite{vpt}, which learns implicit prompts. Our approach is illustrated in Figure~\ref{fig:arch}, which is composed of three basic modules: patch embedding tune, high-frequency components tune as well as Adaptor.

\paragraph{Patch embedding tune.} This module aims at tuning pre-trained patch embedding. In pre-trained SegFormer~\cite{xie2021segformer}, a patch $I^p$ is projected to a $C_{seg}$-dimension feature. We freeze this projection and add a tunable linear layer $\mathtt{L_{pe}}$ to project the original embedding into a $c$-dimension feature $F_{pe} \in \mathbb{R}^c$. 

\begin{align}
	F_{pe} =  \mathtt{L_{pe}}(I^p) \text{, with } c = \frac{C_{seg}}{r} \label{eqn:fpe}
\end{align}
where we introduce the scale factor $r$ to control the tunable parameters.

\paragraph{High-frequency components tune.} For the high frequency components $I_{hfc}$, we learn an overlapped patch embedding similar to SegFormer~\cite{xie2021segformer}. Formally,
$I_{hfc}$ is divided into small patches with the same patch size as SegFormer~\cite{xie2021segformer}. Denoting patch $I^p_{hfc} \in \mathbb{R}^C$ and $C = h \times w \times 3$, we learn a linear layer $\mathtt{L_{hfc}}$ to project the patch into a $c$-dimension feature $F_{hfc} \in \mathbb{R}^c$. 

\begin{align}
	F_{hfc} =  \mathtt{L_{hfc}}(I^p_{hfc})
\end{align}

\paragraph{Adaptor.} The goal of Adaptor is to efficiently and effectively perform adaptation in all the layers by considering features from the image embeddings and high-frequency components.
For the $i$-th Adaptor, we take $F_{pe}$ and $F_{hfc}$ as input and obtain the prompting $P^i$: 
\begin{align}
   	P^i & =  \mathtt{MLP_{up}({GELU(\mathtt{MLP^i_{tune}}(F_{pe}+F_{hfc}))})} 
\end{align}
where $\mathtt{GELU}$ is GELU~\cite{hendrycks2016gaussian} activation. $\mathtt{MLP^i_{tune}}$ is a linear layer for producing different prompts in each Adaptor. $\mathtt{MLP_{up}}$ is an up-projection layer shared across all the Adaptors for matching the dimension of transformer features. $P^i$ is the output prompting that attaches to each transformer layer. 

\section{Experiment}
\label{sec:exp}

\begin{table}
\centering
\resizebox{\columnwidth}{!}
{
\begin{tabular}{c|c|c|c}
\toprule
Task                     & Dataset Name     & \# Train & \# Test \\ \hline
\multirow{2}{*}{\shortstack{Forgery \\Detection}} & CAISA~\cite{dong2013casia}    & 5,123  & 921   \\ \cline{2-4} 
& IMD20~\cite{novozamsky2020imd2020}    & -  & 2,010  \\ \hline
\multirow{2}{*}{\shortstack{Shadow \\Detection}} 
& ISTD~\cite{wang2018stacked}     & 1,330  & 540  \\ \cline{2-4} 
                         & SBU~\cite{sbu}      & 4,089  & 638  \\ \hline
\multirow{2}{*}{\shortstack{Defocus Blur \\Detection}} & CUHK~\cite{shi2014discriminative}     & 604   & 100  \\ \cline{2-4} 
                         &  DUT~\cite{zhao2018defocus}      & -     & 500  \\ \hline
\multirow{3}{*}{ \shortstack{Camouflaged \\ Object Detection}} & COD10K~\cite{fan2020camouflaged}     & 3,040   & 2,026  \\ \cline{2-4} 
& CAMO~\cite{le2019anabranch}    & 1,000     & 250  \\ \cline{2-4}
&  CHAMELEON ~\cite{skurowski2018animal}      & -     & 76  \\ \bottomrule
\end{tabular}
}
\caption{Summary of datasets considered in this work. We show the number of images in training (\textit{\# Train}) and testing set (\textit{\# Test}) for different datasets.}
\label{tab:dataset}
\end{table}

\newcommand{\tablestyle}[2]{\setlength{\tabcolsep}{#1}\renewcommand{\arraystretch}{#2}\centering\footnotesize}

\begin{table*}[t]
	\centering
	\begin{minipage}{0.3\linewidth}
		\centering
        \tablestyle{1pt}{1.05}
        \begin{tabular}{l||cc|cc}
            \toprule
             \multirow{2}{*}{Method} &  \multicolumn{2}{c|}{DUT~\cite{zhao2018defocus}}   & \multicolumn{2}{c}{CUHK~\cite{shi2014discriminative}}  \\  \cline{2-5}
            & $F_{\beta}\uparrow$ & MAE$\downarrow$   & $F_{\beta}\uparrow$ & MAE$\downarrow$    \\ \hline            DeFusionNet~\cite{tang2019defusionnet}  &  .823    & .118  & .818 & .117 \\
            BTBNet~\cite{zhao2019btbnet}  &  .827    & .138  & .889 & .082 \\
            CENet~\cite{zhao2019cenet}    & .817    & .135  & .906 & .059 \\
            DAD~\cite{zhao2021self}     & .794    & .153  & .884    & .079  \\
            EFENet~\cite{zhao2021defocus} & .854    & .094  & .914   & .053  \\ \hline
            Ours & \textbf{.890} & \textbf{.068} & \textbf{.928} & \textbf{.045}  \\ \bottomrule
        \end{tabular} 
        \caption{Comparison with state-of-the-art approaches on defocus blur detection.}
    \label{tab:sota_defocus}
	\end{minipage}\quad
    \vspace{.1em}
	\begin{minipage}{0.3\linewidth}
		\centering
        \tablestyle{1pt}{1.05}
        \begin{tabular}{l||c|c}
            \toprule
             \multirow{2}{*}{Method}  & ISTD~\cite{wang2018stacked} &  SBU~\cite{sbu}  \\   \cline{2-3}
            & BER$\downarrow$ & BER$\downarrow$    \\ \hline
            BDRAR~\cite{zhu2018bidirectional}  & 2.69   & 3.89   \\
            DSC~\cite{hu2018direction}          & 3.42      & 5.59   \\
            DSD~\cite{zheng2019distraction}                 & 2.17 & 3.45   \\
            MTMT~\cite{mtmt}   & 1.72 & 3.15 \\
            FDRNet~\cite{zhu2021mitigating} & 1.55 & \textbf{3.04} \\ \hline
            Ours & \textbf{1.35} & 4.31 \\ \bottomrule
        \end{tabular}
        \caption{Comparison with state-of-the-art approaches on shadow detection.}
    \label{tab:sota_shadow}
	\end{minipage}\quad
    \vspace{.1em}
	\begin{minipage}{0.3\linewidth}
		\centering
        \tablestyle{1pt}{1.05}
        \begin{tabular}{l||cc|cc}
            \toprule
             \multirow{2}{*}{Method} & \multicolumn{2}{c|}{IMD20~\cite{novozamsky2020imd2020}}   & \multicolumn{2}{c}{CAISA~\cite{dong2013casia}}  \\ \cline{2-5}
            & F1$\uparrow$ & AUC$\uparrow$   & F1$\uparrow$ & AUC$\uparrow$      \\ \hline
            ManTra~\cite{wu2019mantra}    & -         & .748    &  -   & .817       \\
            SPAN~\cite{hu2020span}        &    -    & .750         & .382       & .838       \\
            PSCCNet~\cite{liu2022pscc}        &   -  & .806         & .554       & .875       \\
            TransForensics~\cite{hao2021transforensics} & -   & \textbf{.848}    & .627    & .837       \\ 
            ObjectFormer~\cite{wang2022objectformer}    & -   & .821     & .579     & \textbf{.882}       \\\hline
            Ours  & \textbf{.443} & .807 & \textbf{.636} & .862  \\ \bottomrule
        \end{tabular}
        \caption{Comparison with state-of-the-art approaches on forgery detection.}
    \label{tab:sota_forgery}
	\end{minipage}
    \vspace{.1em}
    \vspace{1em}
    \begin{minipage}{1.0\linewidth}
		\centering
        \begin{tabular}{l||cccc|cccc|cccc}
            \toprule
            \multirow{2}{*}{Method}&  \multicolumn{4}{c|}{CHAMELEON~\cite{skurowski2018animal}} & \multicolumn{4}{c|}{CAMO~\cite{le2019anabranch}} & \multicolumn{4}{c}{COD10K~\cite{fan2020camouflaged}}    \\ \cline{2-13}
            & $S_\alpha \uparrow$ & $E_\phi$ $\uparrow$ & $F_\beta^w$ $\uparrow$ & MAE $\downarrow$  & $S_\alpha \uparrow$ & $E_\phi$ $\uparrow$ & $F_\beta^w$ $\uparrow$ & MAE $\downarrow$ & $S_\alpha \uparrow$ & $E_\phi$ $\uparrow$ & $F_\beta^w$ $\uparrow$ & MAE $\downarrow$\\ \hline
            SINet~\cite{fan2020camouflaged} & .869 & .891 & .740 & .044 & .751 & .771 & .606 & .100 & .771 & .806 & .551 & .051 \\
            RankNet~\cite{lv2021simultaneously}  & .846 & .913 & .767 & .045 & .712 & .791 & .583 & .104 & .767 & .861 & .611 & .045 \\
            JCOD~\cite{li2021uncertainty}  & .870 & .924 & - & .039 & .792 & .839 & - & .082 & .800 & .872 & - & .041 \\
            PFNet~\cite{mei2021camouflaged} & .882 & \textbf{.942} & .810 & .033 & .782 & .852 & .695 & .085 & .800 & .868 & .660 & .040  \\
            FBNet~\cite{jiaying2022frequency} & \textbf{.888} & .939 & \textbf{.828} & \textbf{.032} & .783 & .839 & .702 & .081 & .809 & .889 & .684 & .035 \\
            \hline
            Ours & .871 & .917 & .795 & .036 & \textbf{.846} & \textbf{.895} & \textbf{.777} & \textbf{.059} & \textbf{.843} & \textbf{.907} & \textbf{.742} & \textbf{.029} \\
            \bottomrule
        \end{tabular}
        \caption{Comparison with state-of-the-art approaches on camouflaged object detection.}
    \label{tab:sota_cod}
	\end{minipage}\quad
\end{table*}

\begin{table*}[t]
\centering
{
\begin{tabular}{l||c|cc|c|cc|cccc}
\toprule
\multirow{3}{*}{Method}& Trainable & \multicolumn{2}{c|}{\textbf{Defocus Blur}} & \textbf{Shadow} & \multicolumn{2}{c|}{\textbf{Forgery }} & \multicolumn{4}{c}{\textbf{Camouflaged}}\\
      & Param.&  \multicolumn{2}{c|}{CUHK~\cite{shi2014discriminative}} & ISTD~\cite{wang2018stacked}& \multicolumn{2}{c|}{CASIA~\cite{dong2013casia}} & \multicolumn{4}{c}{CAMO~\cite{le2019anabranch}}\\ 
      & (M) &$F_{\beta}\uparrow$ & MAE $\downarrow$      & BER $\downarrow$    & $F_1\uparrow$ & AUC $\uparrow$  & $S_\alpha \uparrow$ & $E_\phi$ $\uparrow$  & $F_\beta^w$ $\uparrow$ & MAE $\downarrow$   \\ \hline
Full-tuning & 64.00 & \textbf{.935} & \textbf{.039} & 2.42  & .465 & .754  & .837 & .887 & \textbf{.778} & .060 \\  
Only Decoder & 3.15 & .891 & .080 & 4.36  & .396 & .722  & .783 & .827 & .671 & .088 \\
VPT-Deep~\cite{vpt} & 3.27 & .913   & .058    & \color{orange}{1.73} & \color{orange}{.588}   & \color{orange}{.847} & .833 & .884 & .751 & .068 \\
AdaptFormer~\cite{chen2022adaptformer} & 3.21 & .912   & .057    & \color{orange}{1.85} & \color{orange}{.602}   & \color{orange}{.855}  & .830 & .877 & .750 & .068 \\
Ours~(r=16) & 3.22 & .924 & .051 & \color{orange}{1.67}  & \color{orange}{.602} & \color{orange}{.857}  & \color{orange}{.838} & \color{orange}{.888} & .761 & .065 \\ 
Ours~(r=4) & 3.70 & .928  & .045  & \color{orange}{\textbf{1.35}} & \color{orange}{\textbf{.636}}   & \color{orange}{\textbf{.862}} & \color{orange}{\textbf{.846}} & \color{orange}{\textbf{.895}} & .777 & \color{orange}{\textbf{.059}} \\ \bottomrule
\end{tabular}}
\caption{
Comparison with state-of-the-art efficient tuning approaches. We conduct evaluations on four datasets for four different tasks.
The efficient tuning method which achieves better performance than full-tuning is marked as {\color{orange}{orange}}.
The best performance among all methods is shown as \textbf{blod}.
}
\label{tab:sota_finetune}
\end{table*}

\subsection{Datasets}
We evaluate our model on a variety of datasets for four tasks: forgery detection, shadow detection, defocus blur detection, and camouflaged object detection. A summary of the basic information of these datasets is illustrated in Table~\ref{tab:dataset}.

\paragraph{Forgery Detection.} CASIA~\cite{dong2013casia} is a large dataset for forgery detection, which is composed of 5,123 training and 921 testing spliced and copy-moved images. 
IMD20~\cite{novozamsky2020imd2020} is a real-life forgery image dataset that consists of 2, 010 samples for testing.
We follow the protocol of previous works~\cite{liu2022pscc, hao2021transforensics,wang2022objectformer} to conduct the training and evaluation at the resolution of $256 \times 256$. We use pixel-level Area Under the Receiver Operating Characteristic Curve~(AUC) and $F_{1}$ score to evaluate the performance. 

\paragraph{Shadow Detection.} SBU~\cite{sbu} is the largest annotated shadow dataset which contains 4,089 training and 638 testing samples, respectively.
ISTD~\cite{wang2018stacked} contains triple samples for shadow detection and removal, we only use the shadowed image and shadow mask to train our method.
Following \cite{mtmt,zhu2018bidirectional,zhu2021mitigating}, we train and test both datasets with the size of $400 \times 400$. 
As for the evaluation metrics, We report the balance error rate~(BER).

\paragraph{Defocus Blur Detection.} 
Following previous work~\cite{zhao2018defocus, cun2020defocus}, we train the defocus blur detection model in the CUHK dataset~\cite{shi2014discriminative}, which contains a total of 704 partial defocus samples. We train the network on the 604 images split from the CUHK dataset and test in DUT~\cite{zhao2018defocus} and the rest of the CUHK dataset. 
The images are resized into $320 \times 320$, following~\cite{cun2020defocus}. We report performances with commonly used metrics: F-measure~($F_{\beta}$) and mean absolute error~(MAE).

\paragraph{Camouflaged Object Detection.} 
COD10K~\cite{fan2020camouflaged} is the largest dataset for camouflaged object detection, which contains 3,040 training and 2,026 testing samples. CHAMELEON~\cite{skurowski2018animal} includes 76 images collected from the Internet for testing. CAMO~\cite{le2019anabranch} provides diverse images with naturally camouflaged objects and artificially camouflaged objects. Following~\cite{fan2020camouflaged,mei2021camouflaged}, we train on the combined dataset and test on the three datasets. We employ commonly used metrics: S-measure~($S_{m}$), mean E-measure~($E_\phi$), weighted F-measure~($F_\beta^w$), and MAE for evaluation.

\subsection{Implementation Details}
All the experiments are performed on a single NVIDIA Titan V GPU with 12G memory. AdamW~\cite{adam} optimizer is used for all the experiments. The initial learning rate is set to $2e^{-4}$ for defocus blur detection and camouflaged object detection, and $5e^{-4}$ for others. Cosine decay is applied to the learning rate. The models are trained for 20 epochs for the SBU~\cite{sbu} dataset and camouflaged combined dataset~\cite{fan2020camouflaged,skurowski2018animal}, and 50 epochs for others. Random horizontal flipping is applied during training for data augmentation. The mini-batch is equal to 4. Binary cross-entropy (BCE) loss is used for defocus blur detection and forgery detection, balanced BCE loss is used for shadow detection, and BCE loss and IOU loss are used for camouflaged object detection. All the experiments are conducted with SegFormer-B4~\cite{xie2021segformer} pre-trained on the ImageNet-1k~\cite{imagenet} dataset.

\begin{figure*}[tp]
    \centering
    \includegraphics[width=0.9\linewidth]{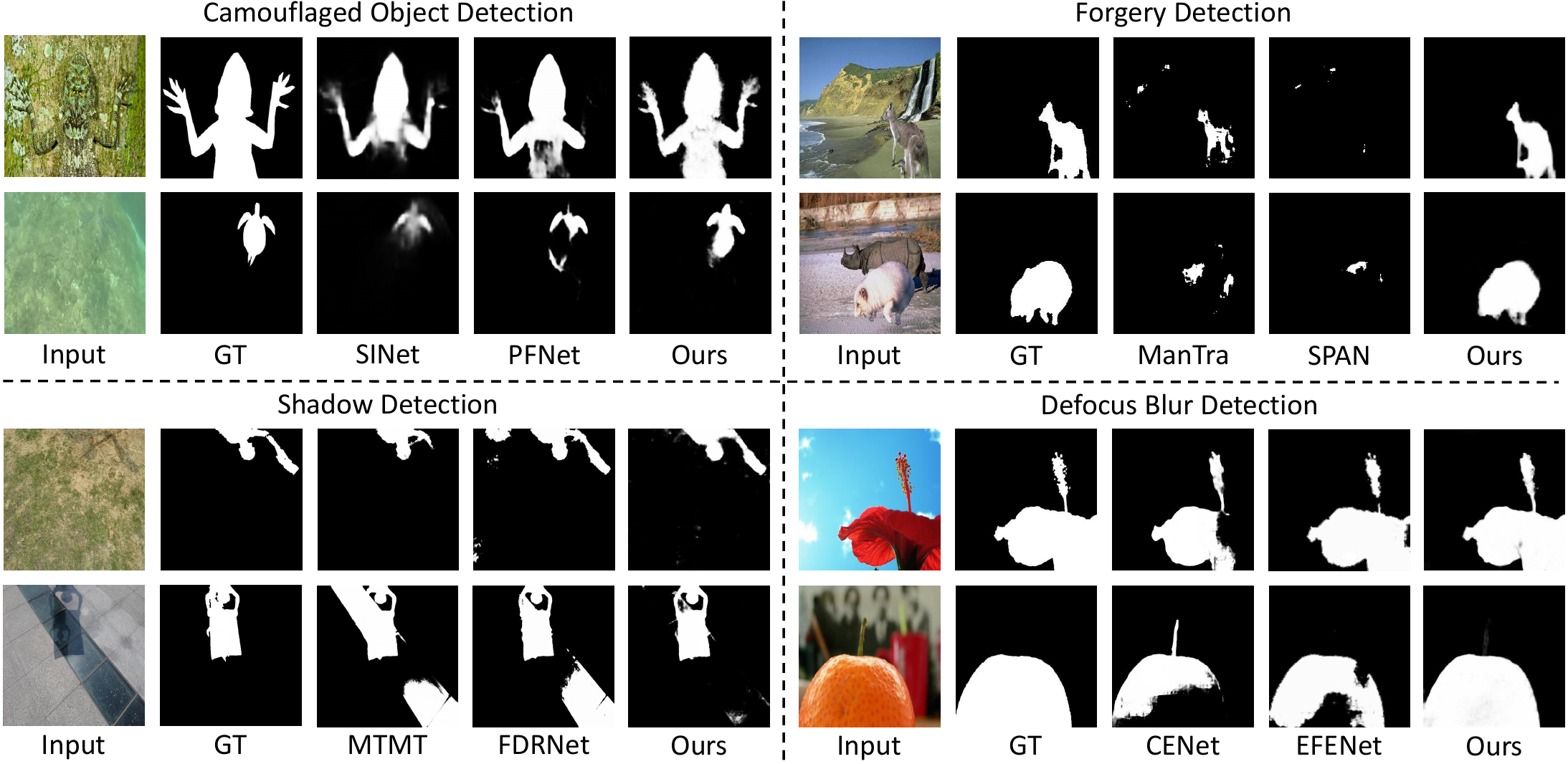}
    \caption{Comparisons with other task-specific methods. We show the results  of: 
    SINet~\cite{fan2020camouflaged} and PFNet~\cite{mei2021camouflaged} on CAMO~\cite{le2019anabranch} dataset for camouflaged object detection (Top-left),
    ManTra~\cite{wu2019mantra} and SPAN~\cite{hu2020span} on CAISA~\cite{dong2013casia} dataset for forgery detection (Top-right),
    MTMT~\cite{mtmt} and FDRNet~\cite{zhu2021mitigating} on ISTD~\cite{wang2018stacked} dataset for shadow detection (Bottom-left), CENet~\cite{zhao2019cenet} and EFENet~\cite{zhao2021defocus} on CUHK~\cite{shi2014discriminative} dataset for defocus blur detection (Bottom-right).}
    \label{fig:sota_result}
\end{figure*}
\begin{table*}[!t]
\centering
{
\begin{tabular}{l||c|cc|c|cc|cccc}
\toprule
\multirow{3}{*}{Method}& Trainable & \multicolumn{2}{c|}{\textbf{Defocus Blur}} & \textbf{Shadow} & \multicolumn{2}{c|}{\textbf{Forgery }} & \multicolumn{4}{c}{\textbf{Camouflaged}}\\
      & Param.&  \multicolumn{2}{c|}{CUHK~\cite{shi2014discriminative}} & ISTD~\cite{wang2018stacked}& \multicolumn{2}{c|}{CASIA~\cite{dong2013casia}} & \multicolumn{4}{c}{CAMO~\cite{le2019anabranch}}\\ 
      & (M) &$F_{\beta}\uparrow$ & MAE $\downarrow$      & BER $\downarrow$    & $F_1\uparrow$ & AUC $\uparrow$  & $S_\alpha \uparrow$ & $E_\phi$ $\uparrow$  & $F_\beta^w$ $\uparrow$ & MAE $\downarrow$   \\ \hline
Decoder~(No prompting) & 3.15 & .891 & .080 & 4.36  & .396 & .722  & .783 & .827 & .671 & .088 \\ 
Ours w/o $F_{pe}$ & 3.61 & .924 & .049 & 1.68  & .540 & .833  & .840 & .887 & .759 & .065 \\ 
Ours w/o $F_{hfc}$ & 3.58 & .926  & .046  & 1.61 & .619   & .846 & .844 & .893 & .773 & .063 \\ 
Ours w/ Shared $\mathtt{MLP^i_{tune}}$ & 3.49 & .928   & .048    & 1.77 & .619   & .860 & .837 & .889 & .763 & .064 \\
Ours w/ Unshared $\mathtt{MLP_{up}}$ & 4.54 & .927   & \textbf{.045}    & \textbf{1.33} & \textbf{.647}   & \textbf{.875}  & .844 & .893 & .774 & .060 \\ 
Ours & 3.70 & \textbf{.928}  & \textbf{.045}  & 1.35 & .636   & .862 & \textbf{.846} & \textbf{.895} & \textbf{.777} & \textbf{.059} \\ \bottomrule
\end{tabular}}
\caption{Ablation on the architecture designs described in Figure~\ref{fig:arch}. We conduct evaluations on four datasets for four different tasks. The proposed prompting strategy (Decoder + $F_{hfc}$ + $F_{pe}$ + Adaptor) performs more effectively.}
\label{tab:arch}
\end{table*}

\begin{table*}[!t]
\centering
{
\begin{tabular}{l||c|cc|c|cc|cccc}
\toprule
Tuning & Trainable & \multicolumn{2}{c|}{\textbf{Defocus Blur}} & \textbf{Shadow} & \multicolumn{2}{c|}{\textbf{Forgery }} & \multicolumn{4}{c}{\textbf{Camouflaged}}\\
      Stage & Param.&  \multicolumn{2}{c|}{CUHK~\cite{shi2014discriminative}} & ISTD~\cite{wang2018stacked}& \multicolumn{2}{c|}{CASIA~\cite{dong2013casia}} & \multicolumn{4}{c}{CAMO~\cite{le2019anabranch}}\\ 
      & (M) &$F_{\beta}\uparrow$ & MAE $\downarrow$      & BER $\downarrow$    & $F_1\uparrow$ & AUC $\uparrow$  & $S_\alpha \uparrow$ & $E_\phi$ $\uparrow$  & $F_\beta^w$ $\uparrow$ & MAE $\downarrow$   \\ \hline
Stage$_{1}$ & 3.16 & .895 & .072 & 3.64  & .408 & .725  & .793 & .834 & .681 & .088 \\ 
Stage$_{1,2}$ & 3.18 & .917 & .058 & 2.45  & .457 & .765  & .806 & .853 & .706 & .081 \\ 
Stage$_{1,2,3}$ & 3.43 & .927   & .047  & 1.46 & .627 & .858 & .841 & .888 & .768 & .062 \\
Stage$_{1,2,3,4}$ & 3.70 & \bf.928  & \bf.045  & \bf1.35 & \bf.636   & \bf.862 & \bf.846 & \bf.895 & \bf.777 & \bf.059 \\ 
\bottomrule
\end{tabular}}
\caption{Ablation on the tuning stages in SegFormer. We conduct evaluations on four datasets for four different tasks. The performance of EVP becomes better as the tuning stages increase.}
\label{tab:tuning_stage}
\end{table*}

\begin{table*}[!t]
\centering
{
\begin{tabular}{c||c|cc|c|cc|cccc}
\toprule
\multirow{3}{*}{$r$}& Trainable & \multicolumn{2}{c|}{\textbf{Defocus Blur}} & \textbf{Shadow} & \multicolumn{2}{c|}{\textbf{Forgery }} & \multicolumn{4}{c}{\textbf{Camouflaged}}\\
      & Param.&  \multicolumn{2}{c|}{CUHK~\cite{shi2014discriminative}} & ISTD~\cite{wang2018stacked}& \multicolumn{2}{c|}{CASIA~\cite{dong2013casia}} & \multicolumn{4}{c}{CAMO~\cite{le2019anabranch}}\\ 
      & (M) &$F_{\beta}\uparrow$ & MAE $\downarrow$      & BER $\downarrow$    & $F_1\uparrow$ & AUC $\uparrow$  & $S_\alpha \uparrow$ & $E_\phi$ $\uparrow$  & $F_\beta^w$ $\uparrow$ & MAE $\downarrow$   \\ \hline
64 & 3.17 & .910 & .055 & 2.09  & .547 & .830  & .829 & .875 & .743 & .070 \\ 
32 & 3.18 & .919 & .054 & 1.84  & .574 & .844  & .832 & .877 & .749 & .067 \\ 
16  & 3.22 & .924  & .051  & 1.67 & .602 & .857 & .838 & .888 & .761 & .065 \\
8  & 3.34 & .923  & .049  & 1.46 & .619 & .856 & .841 & .890 & .767 & .062 \\
4 & 3.70 & .928  & .045   & 1.35 & .636   & \bf.862 & \bf.846 & .895 & .777 & \bf.059 \\
2 & 4.95 & .929  & .042   & \bf1.31 & \bf.642   & .859 & .842 & \bf.896 & .776 & .059 \\ 
1 & 9.56 & \bf.931  & \bf.040   & 1.48 & .621   & .847 & .843 & .894 & \bf.778 & .059 \\ \bottomrule
\end{tabular}}
\caption{Ablation on the parameter scale factor $r$. We conduct evaluations on four datasets for four different tasks. EVP gets the balance between the number of tunable parameters and performances when $r=4$.}
\label{tab:model_size}
\end{table*}

\begin{table*}[t]
\centering
{
\begin{tabular}{l||c|cc|c|cc|cccc}
\toprule
\multirow{3}{*}{Method} & Trainable & \multicolumn{2}{c|}{\textbf{Defocus Blur}} & \textbf{Shadow} & \multicolumn{2}{c|}{\textbf{Forgery }} & \multicolumn{4}{c}{\textbf{Camouflaged}}\\
      & Param.&  \multicolumn{2}{c|}{CUHK~\cite{shi2014discriminative}} & ISTD~\cite{wang2018stacked}& \multicolumn{2}{c|}{CASIA~\cite{dong2013casia}} & \multicolumn{4}{c}{CAMO~\cite{le2019anabranch}}\\ 
      & (M) &$F_{\beta}\uparrow$ & MAE $\downarrow$      & BER $\downarrow$    & $F_1\uparrow$ & AUC $\uparrow$  & $S_\alpha \uparrow$ & $E_\phi$ $\uparrow$  & $F_\beta^w$ $\uparrow$ & MAE $\downarrow$   \\ \hline
Full-tuning & 98.98 & \bf.862 & \bf.077 & 4.39  & .290 & .650  & .593 & \bf.677 & .382 & .157 \\ 
Only Decoder & 13.00 & .836 & .097 & 4.77  & \color{orange}{.318} & \color{orange}{.662}  & \color{orange}{.615} & .659 & \color{orange}{.385} & .162 \\ 
VPT~\cite{vpt}  & 13.09 & .843  & .092  & 4.56 & \color{orange}{.315} & \color{orange}{.666} & \color{orange}{.615} & .660 & \color{orange}{.387} & .161 \\
AdaptFormer~\cite{chen2022adaptformer} & 13.08 & .845   & .092  & 4.60 & \color{orange}{.319} & \color{orange}{.662} & \color{orange}{.614} & .662 & \color{orange}{.387} & .161 \\
EVP & 13.06 & .850  & 087   & \color{orange}{\bf4.36} & \color{orange}{\bf.324}   & \color{orange}{\bf.675} & \color{orange}{\bf.622} & .674 & \color{orange}{\bf.402} & \color{orange}{\bf.156} \\ \bottomrule
\end{tabular}}
\caption{Comparison with other tuning methods with SETR~\cite{zheng2021rethinking} on four different tasks. We conduct an evaluation on four datasets for four different tasks.
The best performance is shown as \textbf{bold}. 
The prompt-tuning method which achieves better performance than full-tuning is marked as \color{orange}{orange}.}
\label{tab:setr}
\end{table*}

\subsection{Main Results}
\label{sec:main_results}

\paragraph{Comparison with the task-specific methods.}
EVP performs well when compared with task-specific methods. We report the comparison of our methods and other task-specific methods in Table~\ref{tab:sota_defocus}, Table~\ref{tab:sota_shadow}, Table~\ref{tab:sota_forgery}, and Table~\ref{tab:sota_cod}. Thanks to our stronger backbone and prompting strategy, EVP achieves the best performance in 5 datasets across 4 different tasks. However, compared with other well-designed domain-specific methods, EVP only introduces a small number of tunable parameters with the frozen backbone and obtains non-trivial performance. We also show some visual comparisons with other methods for each task individually in Figure~\ref{fig:sota_result}. We can see the proposed method predicts more accurate masks compared to other approaches.

\paragraph{Comparison with the efficient tuning methods.}
We evaluate our method with full finetuning and only tuning the decoder, which are the widely-used strategies for down-streaming task adaption. And similar methods from image classification, \ie, VPT~\cite{vpt} and AdaptFormer~\cite{chen2022adaptformer}. 
The number of prompt tokens is set to 10 for VPT and the middle dimension of AdaptMLP is set to 2 for a fair comparison in terms of the tunable parameters.
It can be seen from Table~\ref{tab:sota_finetune} that when only tuning the decoder, the performance drops largely. Compared with similar methods, introducing extra learnable tokens~\cite{vpt} or MLPs in Transformer block~\cite{chen2022adaptformer} also benefits the performance. We introduce a hyper-parameter~($r$) which is used to control the number of parameters of the Adaptor as described in equation~\ref{eqn:fpe}. We first compare EVP~($r$=16) with similar parameters as other methods. From the table, our method achieves much better performance. We also report EVP~($r$=4), with more parameters, the performance can be further improved and outperforms full-tuning on 3 of 4 datasets.

\begin{figure*}[!t]
  \captionsetup[subfigure]{position=b}
  \centering
  \setlength{\tabcolsep}{0pt}
  \subcaptionbox{}{%
    \begin{tabular}{c}
      \includegraphics[width=0.1\textwidth, height=0.1\textwidth]{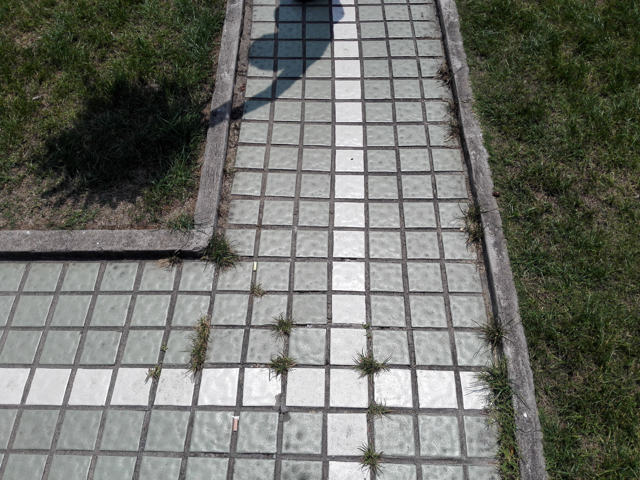} \\[0.1em]
      \includegraphics[width=0.1\textwidth, height=0.1\textwidth]{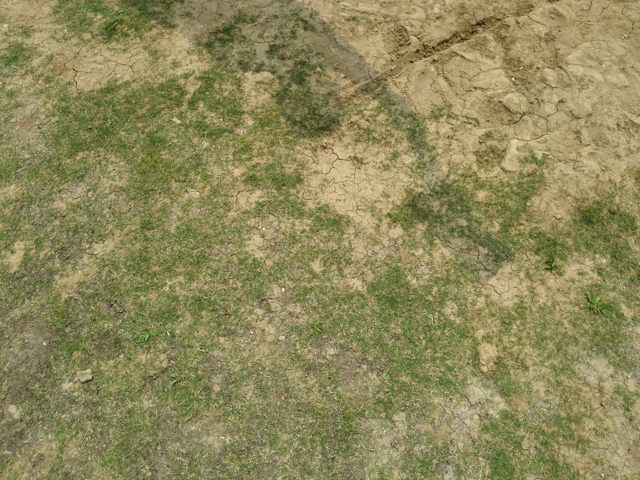} \\[0.1em]
      \includegraphics[width=0.1\textwidth, height=0.1\textwidth]{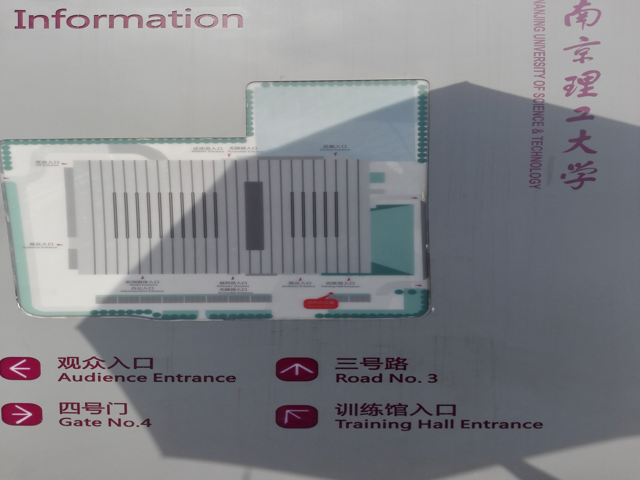} \\
    \end{tabular}}%
  \hspace{0.0em}%
  \subcaptionbox{}{%
    \begin{tabular}{c}
      \includegraphics[width=0.1\textwidth, height=0.1\textwidth]{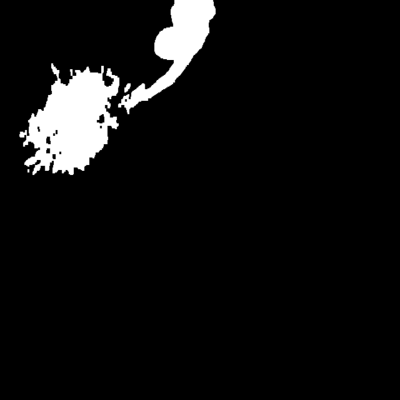} \\[0.1em]
      \includegraphics[width=0.1\textwidth, height=0.1\textwidth]{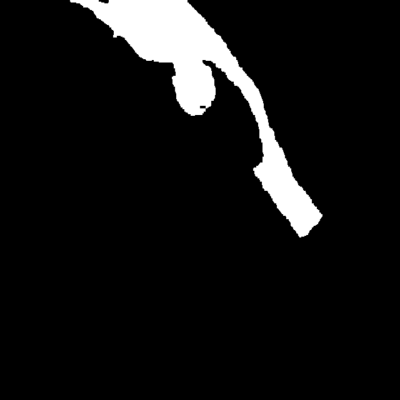} \\[0.1em]
      \includegraphics[width=0.1\textwidth, height=0.1\textwidth]{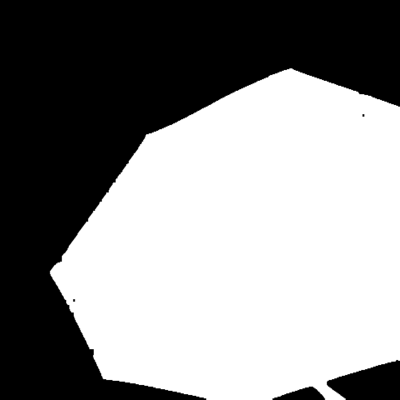} \\
    \end{tabular}}%
  \hspace{0.0em}%
  \subcaptionbox{}{%
    \begin{tabular}{c}
      \includegraphics[width=0.1\textwidth, height=0.1\textwidth]{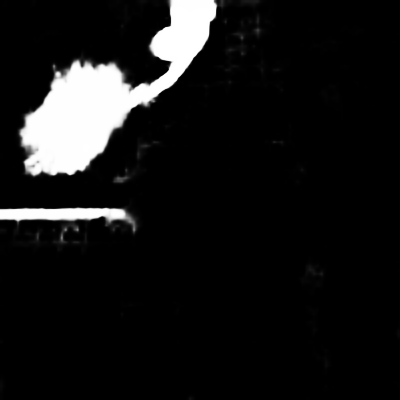} \\[0.1em]
      \includegraphics[width=0.1\textwidth, height=0.1\textwidth]{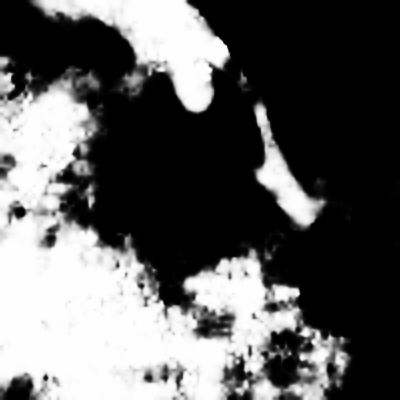} \\[0.1em]
      \includegraphics[width=0.1\textwidth, height=0.1\textwidth]{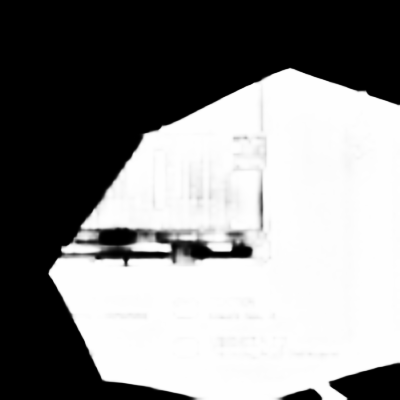} \\
    \end{tabular}}%
  \hspace{0.0em}%
  \subcaptionbox{}{%
    \begin{tabular}{c}
      \includegraphics[width=0.1\textwidth, height=0.1\textwidth]{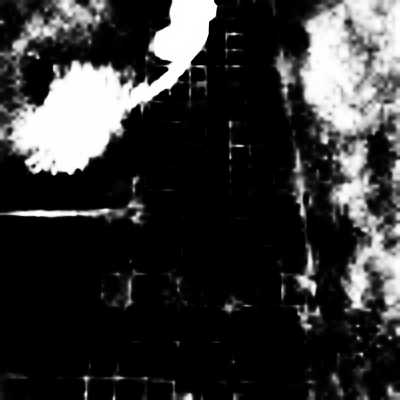} \\[0.1em]
      \includegraphics[width=0.1\textwidth, height=0.1\textwidth]{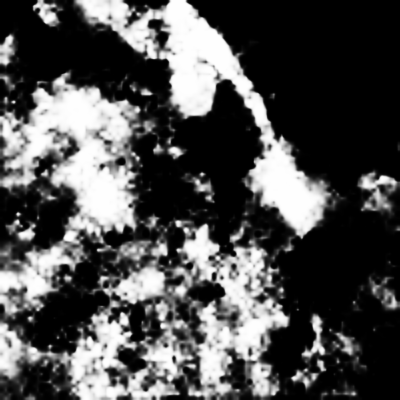} \\[0.1em]
      \includegraphics[width=0.1\textwidth, height=0.1\textwidth]{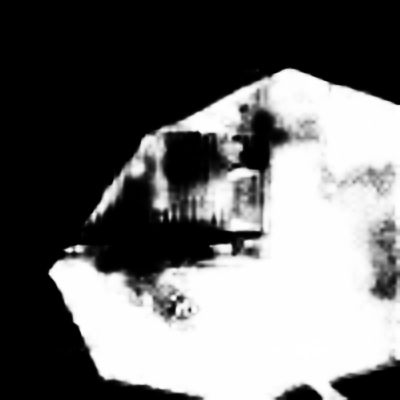} \\
    \end{tabular}}%
  \hspace{0.0em}%
  \subcaptionbox{}{%
    \begin{tabular}{c}
      \includegraphics[width=0.1\textwidth, height=0.1\textwidth]{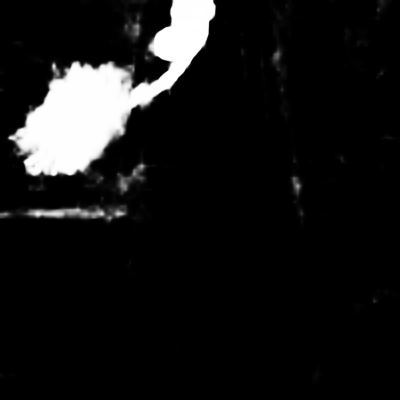} \\[0.1em]
      \includegraphics[width=0.1\textwidth, height=0.1\textwidth]{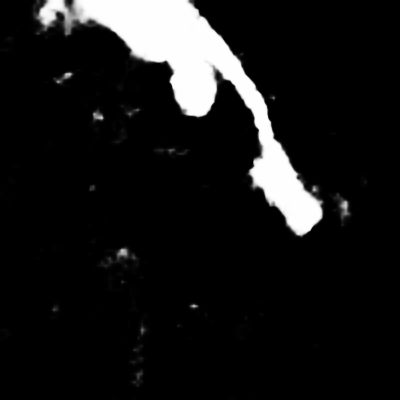} \\[0.1em]
      \includegraphics[width=0.1\textwidth, height=0.1\textwidth]{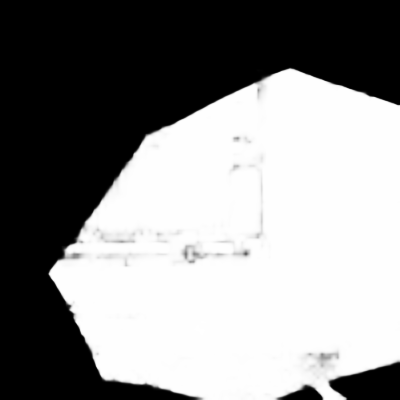} \\
    \end{tabular}}%
  \hspace{0.0em}%
  \subcaptionbox{}{%
    \begin{tabular}{c}
      \includegraphics[width=0.1\textwidth, height=0.1\textwidth]{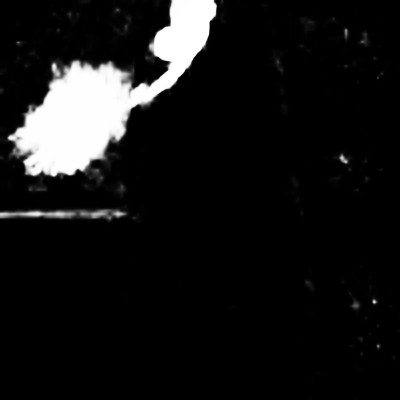} \\[0.1em]
      \includegraphics[width=0.1\textwidth, height=0.1\textwidth]{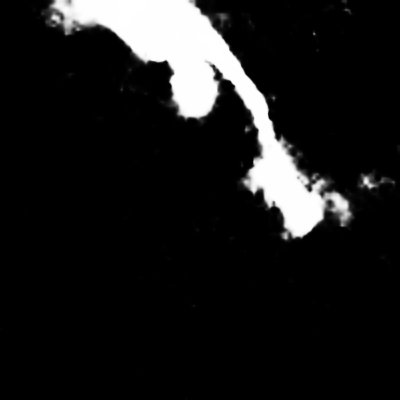} \\[0.1em]
      \includegraphics[width=0.1\textwidth, height=0.1\textwidth]{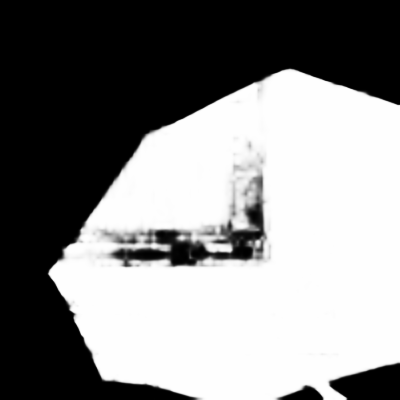} \\
    \end{tabular}}%
  \hspace{0.0em}%
    \subcaptionbox{}{%
    \begin{tabular}{c}
      \includegraphics[width=0.1\textwidth, height=0.1\textwidth]{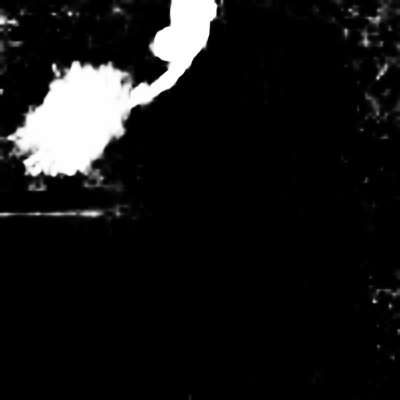} \\[0.1em]
      \includegraphics[width=0.1\textwidth, height=0.1\textwidth]{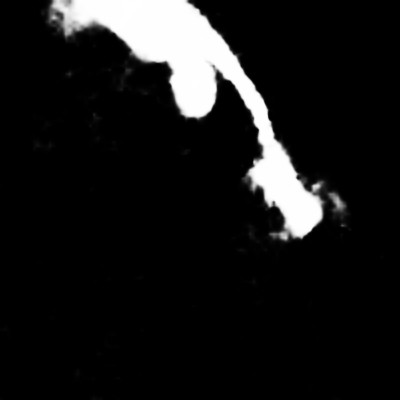} \\[0.1em]
      \includegraphics[width=0.1\textwidth, height=0.1\textwidth]{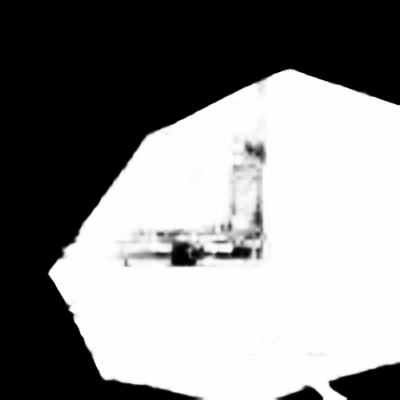} \\
    \end{tabular}}%
  \hspace{0.0em}%
  \subcaptionbox{}{%
    \begin{tabular}{c}
      \includegraphics[width=0.1\textwidth, height=0.1\textwidth]{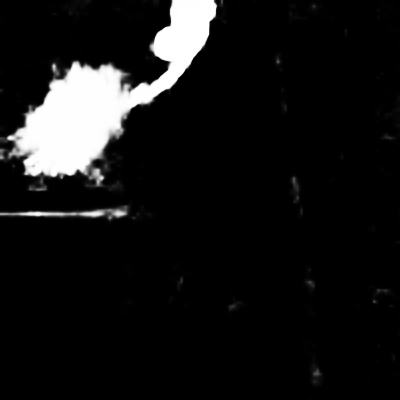} \\[0.1em]
      \includegraphics[width=0.1\textwidth, height=0.1\textwidth]{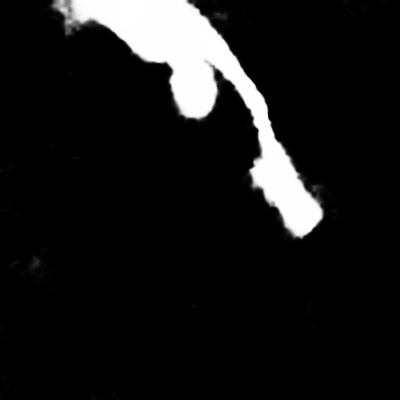} \\[0.1em]
      \includegraphics[width=0.1\textwidth, height=0.1\textwidth]{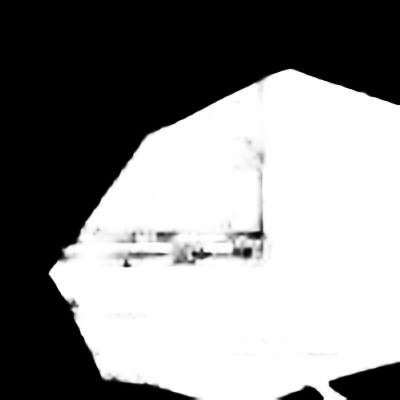} \\
    \end{tabular}}%
  \hspace{0.0em}%
  \subcaptionbox{}{%
    \begin{tabular}{c}
      \includegraphics[width=0.1\textwidth, height=0.1\textwidth]{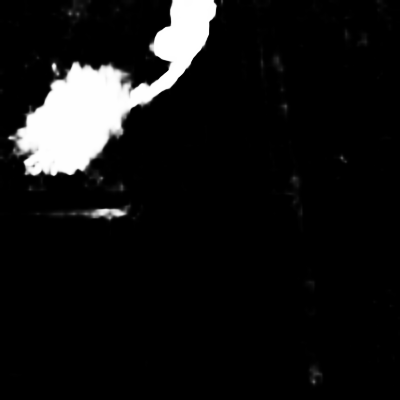} \\[0.1em]
      \includegraphics[width=0.1\textwidth, height=0.1\textwidth]{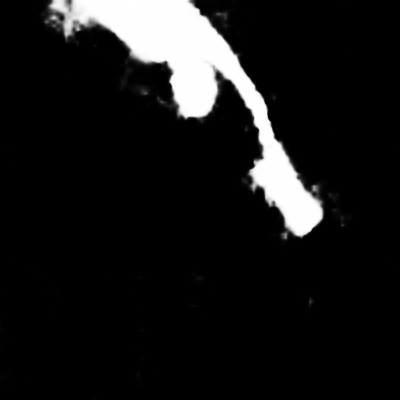} \\[0.1em]
      \includegraphics[width=0.1\textwidth, height=0.1\textwidth]{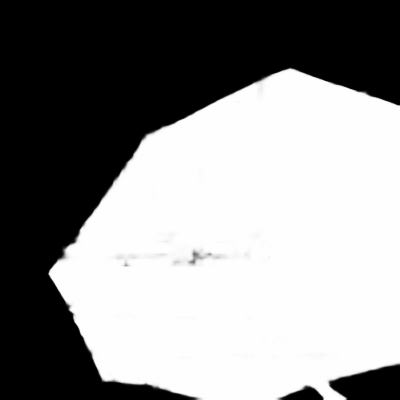} \\
    \end{tabular}}%
  \hspace{0.0em}%
\caption{Quantitative comparison using full-tuning and different prompting designs on ISTD~\cite{wang2018stacked} dataset for shadow detection. From the left to right is: (a)~Input, (b)~GT, (c)~Full-tuning, (d)~Decoder~(No prompting), (e)~Ours w/o $F_{pe}$, (f)~Ours w/o $F_{hfc}$, (g)~Ours w/ Shared $\mathtt{MLP^i_{tune}}$, (h)~Ours w/ Unshared $\mathtt{MLP_{up}}$, (i)~Ours Full.}
\label{fig:ablation_arch}
\end{figure*}

\subsection{Ablation Study}
\label{sec:ablation_study}
We conduct the ablation to show the effectiveness of each component. The experiments are performed with the scaling factor $r=4$ except specified.

\paragraph{Architecture Design.}
To verify the effectiveness of the proposed visual prompting architecture, we modify it into different variants. As shown in Table~\ref{tab:arch} and Figure~\ref{fig:ablation_arch}, sharing $\mathtt{MLP^i_{tune}}$ in different Adaptors only saves a small number of parameters~(0.55M \textit{v.s.} 0.34M) but leads to a significant performance drop. It cannot obtain consistent performance improvement when using different $\mathtt{MLP_{up}}$ in different Adaptors, moreover introducing a large number of parameters~(0.55M \textit{v.s.} 1.39M). On the other hand, the performance will drop when we remove $F_{pe}$ or $F_{hfc}$, which  means that they are both effective visual prompts.

\paragraph{Tuning Stage.}
We try to answer the question: which stage contributes mostly to prompting tuning? Thus, we show the variants of our tuning method by changing the tunable stages in the SegFormer backbone. SegFormer contains 4 stages for multi-scale feature extraction. We mark the Stage$_{x}$ where the tunable prompting is added in Stage $x$. Table~\ref{tab:tuning_stage} shows that better performance can be obtained via the tunable stages increasing. Besides, the maximum improvement occurs in Stage$_{1,2}$ to Stage$_{1,2,3}$. Note that the number of transformer blocks of each stage in SegFormer-B4 is 3, 8, 27, and 3, respectively. Thus, the effect of EVP is positively correlated to the number of the prompted transformer blocks.

\paragraph{Scale Factor $r$~(equation \ref{eqn:fpe}).}
We introduce $r$ in Sec~\ref{sec:explicit_visual_prompting} of the main paper to control the number of learnable parameters. A larger $r$ will use fewer parameters for tuning. As shown in Table~\ref{tab:model_size}, the performance improves on several tasks when $r$ decreases from 64 to 4; when $r$ continues to decrease to 2 or 1, it can not gain better performance consistently even if the model becomes larger. It indicates that $r=4$ is a reasonable choice to make a trade-off between the performance and model size.

\paragraph{EVP in Plain ViT.}
We experiment on SETR~\cite{zheng2021rethinking} to confirm the generalizability of EVP. SETR employs plain ViT as the backbone and a progressive upsampling ConvNet as the decoder, while SegFormer has a hierarchical backbone with 4 stages. Therefore, the only distinction between the experiments using SegFormer is that all modifications are limited to the single stage in plain ViT. The experiments are conducted with ViT-Base~\cite{dosovitskiy2020image} pretrained on the ImageNet-21k~\cite{imagenet} dataset. The number of prompt tokens is set to 10 for VPT, the middle dimension of AdaptMLP is set to 4 for AdaptFormer, and $r$ is set to 32 for our EVP. As shown in Table~\ref{tab:setr}, EVP also outperforms other tuning methods when using plain ViT as the backbone.

\section{Conclusion}
\label{sec:conclusion}

In this paper, we present explicit visual prompting to unify the solutions of low-level structure segmentations.
We mainly focus on two kinds of features: the frozen features from patch embedding and the high-frequency components from the original image. Equipped with our method, we find that a frozen vision transformer backbone from the ImageNet with limited tunable parameters can achieve similar performance as the full-tuned network structures, also the state-of-the-art performance compared with the other task-specific methods. For future works, we will extend our approach to other related problems and hope it can promote further exploration of visual prompting.

\paragraph{Acknowledgments.}
\label{sec:acknowledgments}
This work was supported in part by the University of Macau under Grant MYRG2022-00190-FST and in part by the Science and Technology Development Fund, Macau SAR,
under Grants: 0034/2019/AMJ, 0087/2020/A2, and 0049/2021/A.

{\small
\bibliographystyle{ieee_fullname}
\bibliography{11_references}
}

\ifarxiv \clearpage \appendix
\label{sec:appendix}
\renewcommand\thetable{\Alph{section}\arabic{table}}   
\renewcommand\thefigure{\Alph{section}\arabic{figure}}  

\section{Implementation Details}
\label{sec:implementation_details}
We give more implementation details in the main paper of the "comparison with the task-specific methods" and "comparison with the efficient tuning methods". 

\paragraph{Basic Setting.} Our method contains a backbone for feature extraction and a decoder for segmentation prediction. We initialize the weight of the backbone via ImageNet classification pre-training, and the weight of the decoder is randomly initialized. Below, we give the details of each variant.

\paragraph{Full-tuning.} We follow the basic setting above, and then, fine-tune all the parameters of the encoder and decoder.

\paragraph{Only Decoder.} We follow the basic setting above, and then, fine-tune the parameters in the decoder only.

\paragraph{VPT~\cite{vpt}.} We first initialize the model following the basic setting. Then, we concatenate the prompt embeddings in each transformer block of the backbone only. Notice that, their prompt embeddings are implicitly shared across the whole dataset. We follow their  original paper and optimize the parameters in the prompt embeddings and the decoder. 

\paragraph{AdaptFormer~\cite{chen2022adaptformer}.} We first initialize the model following the basic setting above. Then, the AdaptMLP is added to each transformer block of the backbone for feature adaptation. We fine-tune the parameters in the decoder and the newly introduced AdaptMLP. 

\paragraph{EVP~(Ours).} We also initialize the weight following the basic setting. Then, we add the explicit prompting as described in the main paper of Figure~\ref{fig:arch}. 

\paragraph{Metric.}
AUC calculates the area of the ROC curve. ROC curve is a function of true positive rate~($\frac{tp}{tp + fn}$) in terms of false positive rate~($\frac{fp}{fp + tn}$), where $tp$, $tn$, $fp$, $fn$ represent the number of pixels which are classified as true positive, true negative, false positive, and false negative, respectively. $F_{1}$ score is defined as $F_{1} = \frac{2 \times precision \times recall}{precision + recall}$, where $precision = \frac{tp}{tp + fp}$ and $recall = \frac{tp}{tp + fn}$. 
The balance error rate~(BER) $ = \left(1-\frac{1}{2}\left(\frac{tp}{tp+fn}+\frac{tn}{tn+fp}\right)\right) \times 100$.
F-measure is calculated as $F_{\beta} = \frac{\left(1+\beta^2\right) \times precision \times recall}{\beta^2 \times precision + recall}$, where $\beta^2=0.3$. MAE computes pixel-wise average distance.
Weighted F-measure ($F_\beta^w$) weighting the quantities TP, TN, FP, and FN according to the errors at their location and their neighborhood information:
$F_\beta^w = \frac{\left(1+\beta^2\right) \times precision^w \times recall^w}{\beta^2 \times precision^w + recall^w}$.
E-measure~($E_\phi$) jointly considers image statistics and local pixel matching:
$E_\phi=\frac{1}{W \times H} \sum_{i=1}^W \sum_{j=1}^H \phi_S(i, j)$,
where $\phi_S$ is the alignment matrix depending on the similarity of the prediction and ground truth.

\paragraph{Training Data.}
Note that most forgery detection methods~(ManTraNet~\cite{wu2019mantra}, SPAN~\cite{hu2020span}, PSCCNet~\cite{liu2022pscc}, and ObjectFormer~\cite{wang2022objectformer} in Table~\ref{tab:sota_forgery}) and one shadow detection method~(MTMT~\cite{mtmt} in Table~\ref{tab:sota_shadow}) use extra training data to get better performance. We only use the training data from the standard datasets and obtain SOTA performance.

\section{More Results}
\label{sec:more_results}
We provide more experimental results in addition to the main paper.

\subsection{High-Frequency Prompting}
Our method gets the knowledge from the explicit content of the image itself, hence we also discuss other similar explicit clues of images as the prompts. Specifically, we choose the common-used Gaussian filter, the noise-filter~\cite{fridrich2012rich}, the all-zero image, and the original image as experiments. From Table~\ref{tab:filter}, we find the Gaussian filter shows a better performance in defocus blur since 
it is also a kind of blur. Also, the noise filter~\cite{fridrich2012rich} from forgery detection also boosts the performance. Interestingly, we find that simply replacing the original image with an all-zero image also boosts the performance, since it can also be considered as a kind of implicitly learned embeddings across the full dataset as in VPT~\cite{vpt}. Differently, the high-frequency components of the image achieve consistent performance improvement to other methods on these several benchmarks.

\subsection{HFC \textit{v.s.} LFC} We conduct the ablation study on choosing of high-frequency features or the low-frequency features in Table~\ref{tab:hfc}. From the table, using the low-frequency components as the prompting just show some trivial improvement on these datasets. Differently, the high-frequency components are more general solutions and show a much better performance in shadow detection, forgery detection, and camouflaged detection. Similar to the Gaussian filter as we discussed above, the LFC is also a kind of blur, which makes the advantage of LFC in the defocus blur detection.

\subsection{Mask Ratio $\tau$} We further evaluate the hyper-parameter mask ratio~$\tau$ introduced in Section~\ref{Preliminary}. From Table~\ref{tab:hfc}, when we mask out 25\% of the central pixels in the spectrum, it achieves consistently better performance in all the tasks. We also find that the performance may drop when the increasing of mask ratio~(all 0 images), especially in shadow detection, forgery detection, and camouflaged object detection.

\begin{table*}[t]
\centering
\begin{tabular}{l|cc|c|cc|cccc}
\toprule
\multirow{3}{*}{Method}& \multicolumn{2}{c|}{\textbf{Defocus Blur}} & \textbf{Shadow} & \multicolumn{2}{c|}{\textbf{Forgery}}  & \multicolumn{4}{c}{\textbf{Camouflaged}}\\
&  \multicolumn{2}{c|}{CUHK~\cite{shi2014discriminative}} & ISTD~\cite{wang2018stacked} & \multicolumn{2}{c|}{CASIA~\cite{dong2013casia}}& \multicolumn{4}{c}{CAMO~\cite{le2019anabranch}} \\ 
&$F_{\beta}\uparrow$ & MAE $\downarrow$      & BER $\downarrow$    & $F_1\uparrow$ & AUC $\uparrow$  & $S_\alpha \uparrow$ & $E_\phi$ $\uparrow$  \\ \hline
Gaussian Blur & \bf.928  & .046   & 1.52  & .631  & .860 & .842 & .893   \\ 
Noise Filter  & .923 & .046   & 1.47 & \bf.637  & \bf.866 & .843 & .894   \\
All 0 Image   & .922 & .047  & 1.45 & .630 & .862  & .844 & .895 \\ 
Original  & .922 & .047  & 1.59 & .630 & .860  & .840 & .891  \\ 
\rowcolor{lightgray!30} HFC  &\bf .928 & \bf.045  & \bf 1.35 &  .636 &  .862  & \bf.846 & \bf.895  \\
\bottomrule
\end{tabular}
\caption{Ablation on the explicit visual prompting with other relatives. We compare with the widely-used image filter as prompting to verify the effectiveness of the proposed HFC.
}
\label{tab:filter}
\end{table*}

\begin{table*}[t]
\centering
\begin{tabular}{lc|cc|c|cc|cccc}
\toprule
\multirow{3}{*}{Method}& Mask & \multicolumn{2}{c|}{\textbf{Defocus Blur}} & \textbf{Shadow} & \multicolumn{2}{c|}{\textbf{Forgery}}  & \multicolumn{4}{c}{\textbf{Camouflaged}}\\
      & Ratio &  \multicolumn{2}{c|}{CUHK~\cite{shi2014discriminative}} & ISTD~\cite{wang2018stacked}& \multicolumn{2}{c|}{CASIA~\cite{dong2013casia}}& \multicolumn{4}{c}{CAMO~\cite{le2019anabranch}} \\ 
      & $\tau$ (\%) &$F_{\beta}\uparrow$ & MAE $\downarrow$      & BER $\downarrow$    & $F_1\uparrow$ & AUC $\uparrow$  & $S_\alpha \uparrow$ & $E_\phi$ $\uparrow$  \\ \hline
\multicolumn{11}{c}{\textbf{Low Frequency Components (LFC) with FFT}} \\
LFC* & 0 & .922 & .047  & 1.45 & .630 & .862  & .844 & .895 \\ 
LFC         & 10 & .927 & .046  & 1.58 & .631 & .862  & .845 & \bf.895 \\
LFC         & 25 & .924 & .047 & 1.49 & .630 & .860  & .842 & .891 \\
LFC         & 50 & .923 & .048  & 1.48 & .630 & .860  & .841 & .893 \\
LFC         & 75 & .924 & 048  & 1.56 & .627 & .859  & .840 & .894 \\
LFC         & 90 & .925 & .046  & 1.47 & .626 & .859  & .841 & .894 \\
LFC** & 100  & .922 & .047  & 1.59 & .630 & .860  & .840 & .891  \\ \hline
\multicolumn{11}{c}{\textbf{High Frequency Components (HFC) with FFT}} \\
HFC & 10 & .926 & .046  &  1.60 &  .631 & .854  & .843 & .894 \\
\rowcolor{lightgray!30} HFC & 25 &\bf .928 & \bf.045  & \bf 1.35 &  \bf .636 &  \bf .862  & \bf.846 & \bf.895  \\
HFC & 50 & .925 & .047  & 1.51 & .631 & .862  & .843 & .894  \\
HFC & 75 & .923 & .048 & 1.52 & .629 & .858  & .842 & .892  \\
HFC & 90 & .924 & .047  & 1.49 & .630 & .861  & .842 & .893\\
\bottomrule
\end{tabular}
\caption{Ablation on HFC, LFC, and mask ratio $\tau$. Leveraging FFT to extract high-frequency components consistently outperforms LFC. *LFC~($\tau$=0) equals to a full zero image for prompting. **LFC~($\tau$=100) means we learn an embedding from the original input image directly. }
\vspace{-1em}
\label{tab:hfc}
\end{table*}

\section{Additional Visual Results}
\label{sec:visulation}

We give more visual results of EVP and other task-specific methods on the four tasks in Figure~\ref{fig:supp_forgery},~\ref{fig:supp_shadow},~\ref{fig:supp_defocus}, and~\ref{fig:supp_cod} as supplementary to the visual results in the main paper.

\begin{figure*}[t]
  \captionsetup[subfigure]{position=b}
  \centering

  \setlength{\tabcolsep}{0pt}
  \subcaptionbox{\scriptsize{Input}}{%
    \begin{tabular}{c}
      \includegraphics[width=0.19\textwidth, height=0.15\textwidth]{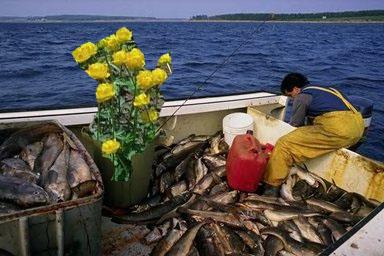} \\[0.1em]
      \includegraphics[width=0.19\textwidth, height=0.15\textwidth]{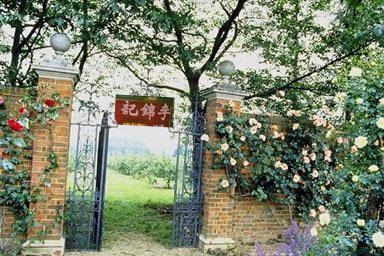} \\[0.1em]
      \includegraphics[width=0.19\textwidth, height=0.15\textwidth]{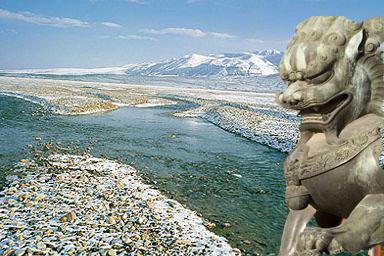}\\[0.1em]
      \includegraphics[width=0.19\textwidth, height=0.15\textwidth]{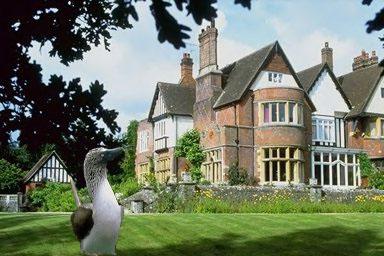}\\[0.1em]
      \includegraphics[width=0.19\textwidth, height=0.15\textwidth]{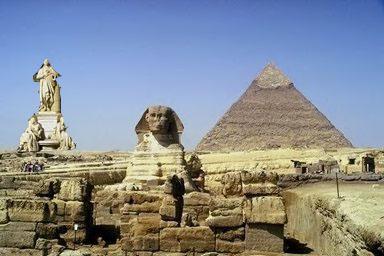}
    \end{tabular}}%
  \hspace{0.1em}%
  \subcaptionbox{\scriptsize{GT}}{%
    \begin{tabular}{c}
      \includegraphics[width=0.19\textwidth, height=0.15\textwidth]{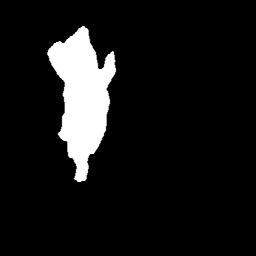}\\[0.1em]
      \includegraphics[width=0.19\textwidth, height=0.15\textwidth]{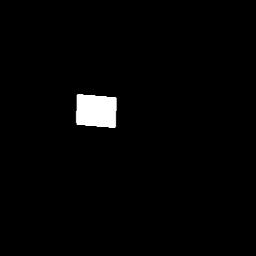} \\[0.1em]
      \includegraphics[width=0.19\textwidth, height=0.15\textwidth]{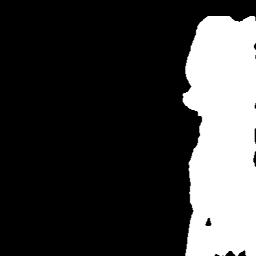}\\[0.1em]
      \includegraphics[width=0.19\textwidth, height=0.15\textwidth]{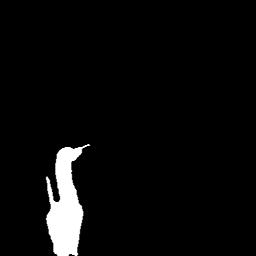}\\[0.1em]
      \includegraphics[width=0.19\textwidth, height=0.15\textwidth]{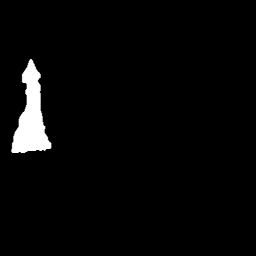}
    \end{tabular}}%
  \hspace{0.1em}%
  \subcaptionbox{\scriptsize{Ours}}{%
    \begin{tabular}{c}
      \includegraphics[width=0.19\textwidth, height=0.15\textwidth]{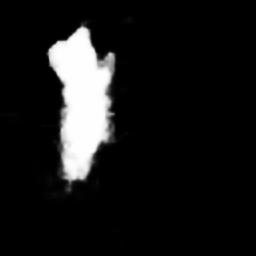} \\[0.1em]
      \includegraphics[width=0.19\textwidth, height=0.15\textwidth]{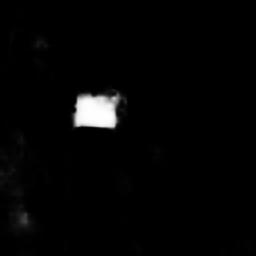}\\[0.1em]
      \includegraphics[width=0.19\textwidth, height=0.15\textwidth]{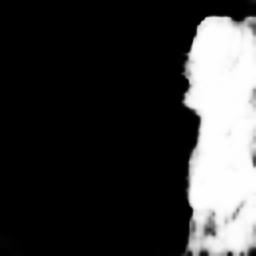}\\[0.1em]
      \includegraphics[width=0.19\textwidth, height=0.15\textwidth]{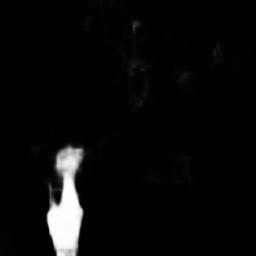}\\[0.1em]
      \includegraphics[width=0.19\textwidth, height=0.15\textwidth]{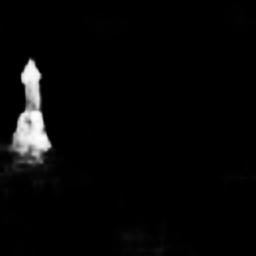}
    \end{tabular}}%
  \hspace{0.1em}%
  \subcaptionbox{\scriptsize{ManTraNet}}{%
    \begin{tabular}{c}
      \includegraphics[width=0.19\textwidth, height=0.15\textwidth]{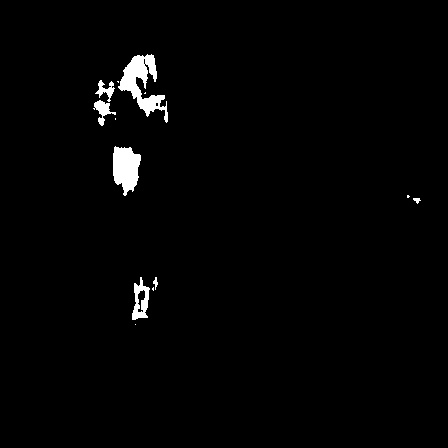} \\[0.1em]
      \includegraphics[width=0.19\textwidth, height=0.15\textwidth]{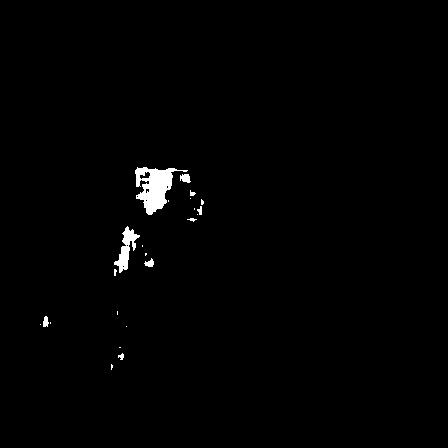} \\[0.1em]
      \includegraphics[width=0.19\textwidth, height=0.15\textwidth]{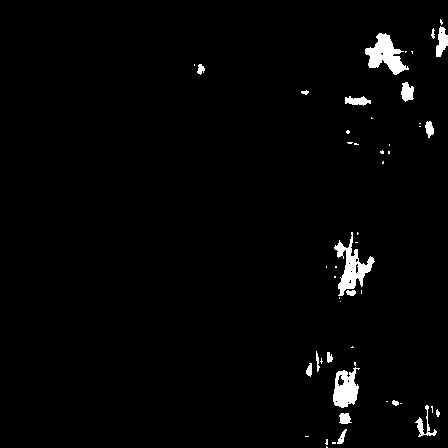}\\[0.1em]
      \includegraphics[width=0.19\textwidth, height=0.15\textwidth]{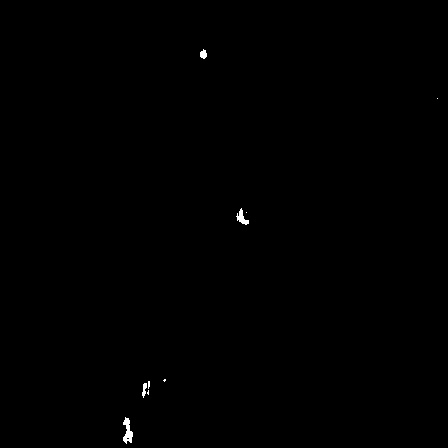}\\[0.1em]
      \includegraphics[width=0.19\textwidth, height=0.15\textwidth]{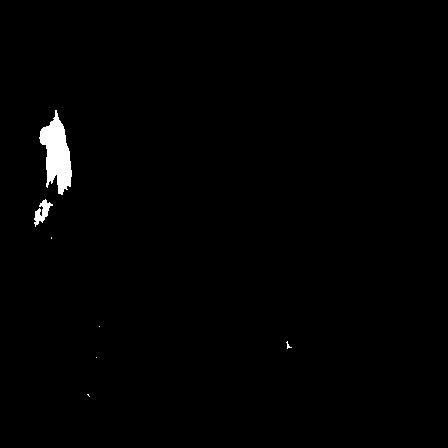}
    \end{tabular}}%
  \hspace{0.1em}%
  \subcaptionbox{\scriptsize{SPAN}}{%
    \begin{tabular}{c}
      \includegraphics[width=0.19\textwidth, height=0.15\textwidth]{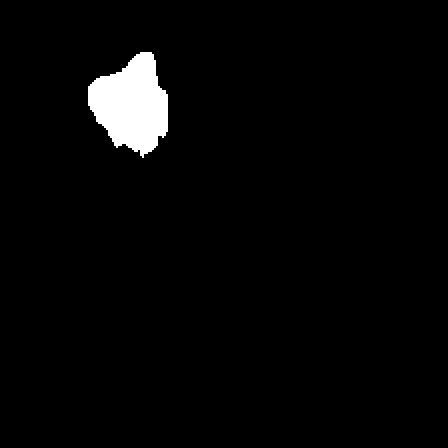}\\[0.1em]
      \includegraphics[width=0.19\textwidth, height=0.15\textwidth]{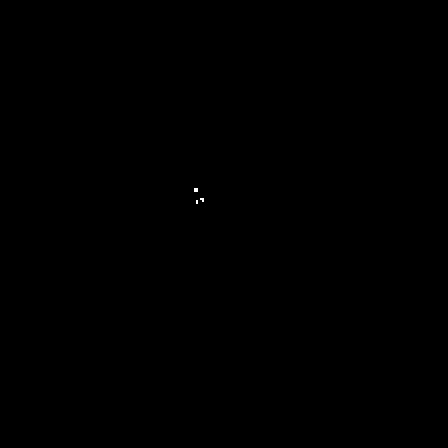} \\[0.1em]
      \includegraphics[width=0.19\textwidth, height=0.15\textwidth]{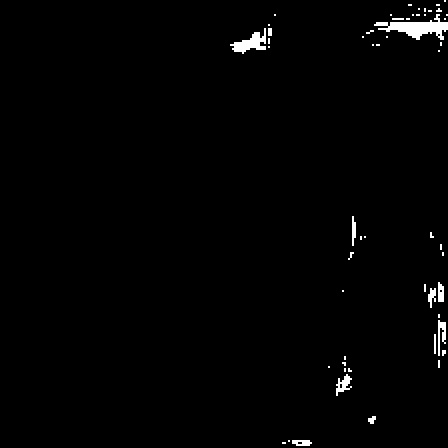}\\[0.1em]
      \includegraphics[width=0.19\textwidth, height=0.15\textwidth]{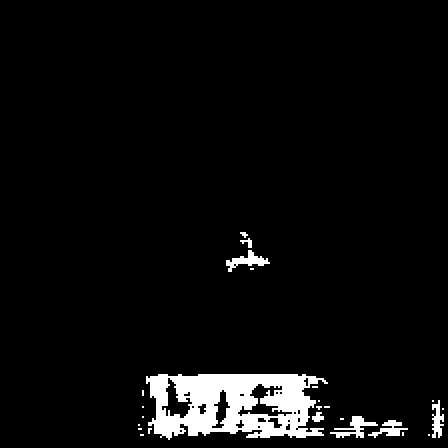}\\[0.1em]
      \includegraphics[width=0.19\textwidth, height=0.15\textwidth]{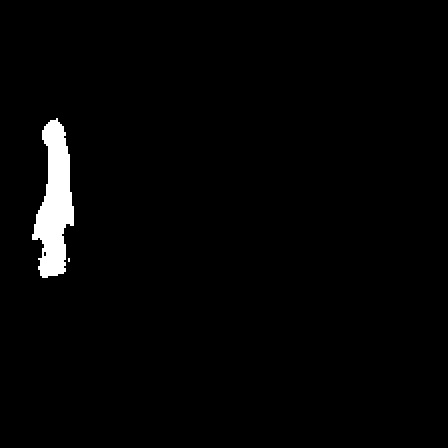}
    \end{tabular}}%
  \hspace{0.0em}%
  
\caption{More results on CAISA~\cite{dong2013casia} dataset for forgery detection. We compare to ManTraNet~\cite{wu2019mantra} and SPAN~\cite{hu2020span}.}
\label{fig:supp_forgery}
\end{figure*}

\begin{figure*}[t]
  \captionsetup[subfigure]{position=b}
  \centering
  \setlength{\tabcolsep}{0pt}
  \subcaptionbox{\scriptsize{Input}}{%
    \begin{tabular}{c}
      \includegraphics[width=0.15\textwidth, height=0.15\textwidth]{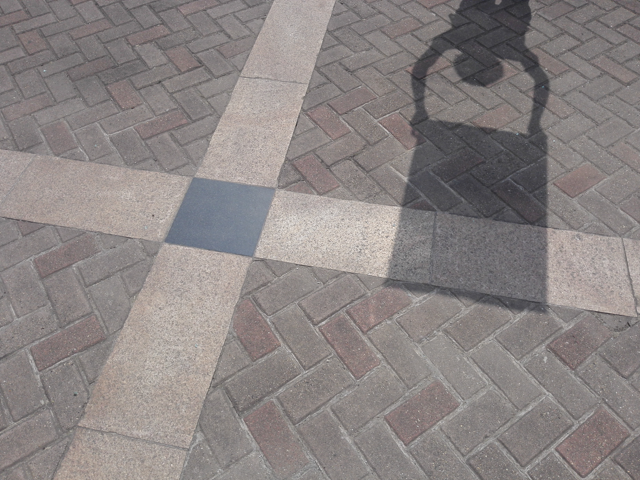} \\[0.1em]
      \includegraphics[width=0.15\textwidth, height=0.15\textwidth]{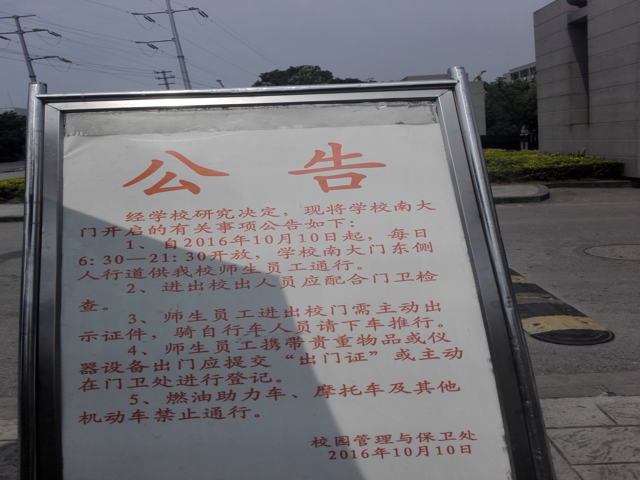} \\[0.1em]
      \includegraphics[width=0.15\textwidth, height=0.15\textwidth]{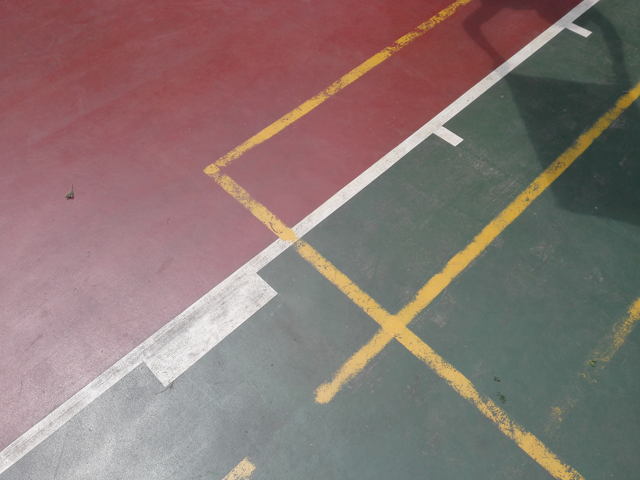}\\[0.1em]
      \includegraphics[width=0.15\textwidth, height=0.15\textwidth]{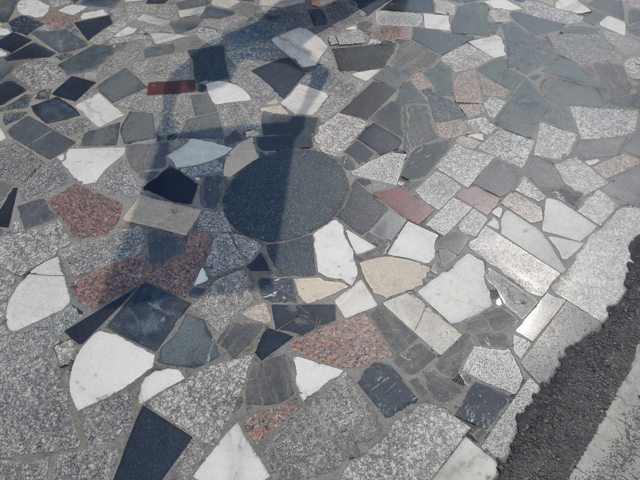}\\[0.1em]
      \includegraphics[width=0.15\textwidth, height=0.15\textwidth]{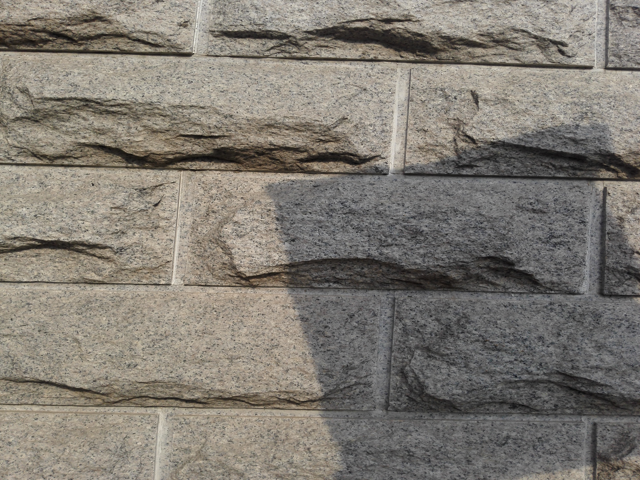}
    \end{tabular}}%
  \hspace{0.1em}%
  \subcaptionbox{\scriptsize{GT}}{%
    \begin{tabular}{c}
      \includegraphics[width=0.15\textwidth, height=0.15\textwidth]{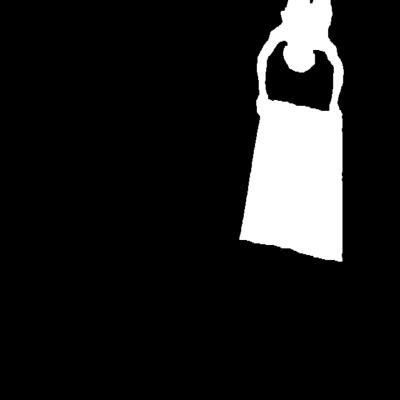} \\[0.1em]
      \includegraphics[width=0.15\textwidth, height=0.15\textwidth]{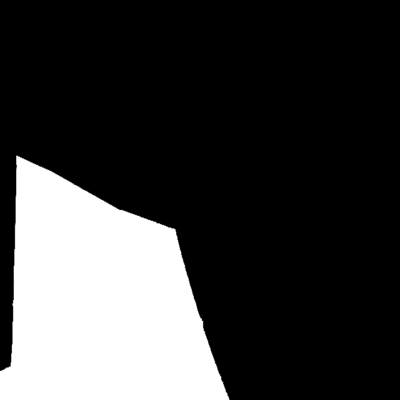} \\[0.1em]
      \includegraphics[width=0.15\textwidth, height=0.15\textwidth]{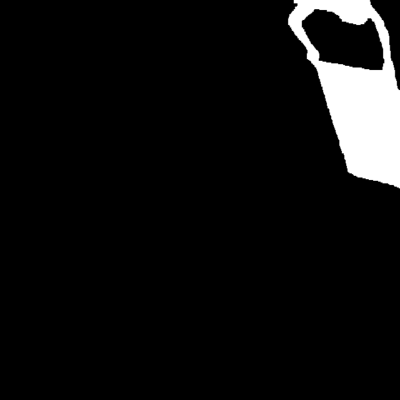}\\[0.1em]
      \includegraphics[width=0.15\textwidth, height=0.15\textwidth]{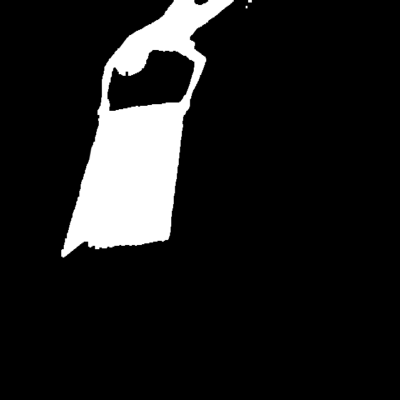}\\[0.1em]
      \includegraphics[width=0.15\textwidth, height=0.15\textwidth]{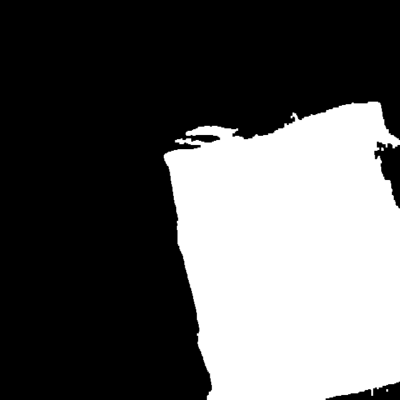}
    \end{tabular}}%
  \hspace{0.1em}%
  \subcaptionbox{\scriptsize{Ours}}{%
    \begin{tabular}{c}
      \includegraphics[width=0.15\textwidth, height=0.15\textwidth]{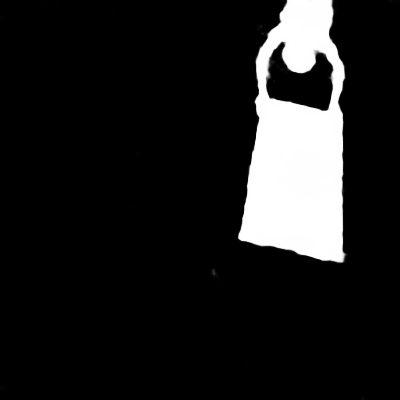} \\[0.1em]
      \includegraphics[width=0.15\textwidth, height=0.15\textwidth]{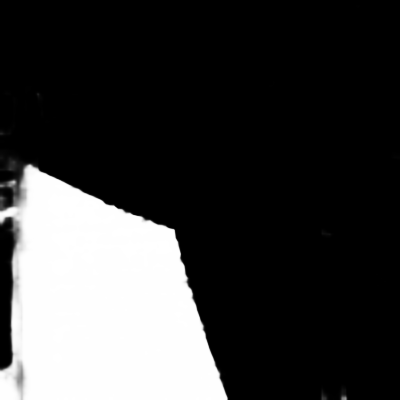} \\[0.1em]
      \includegraphics[width=0.15\textwidth, height=0.15\textwidth]{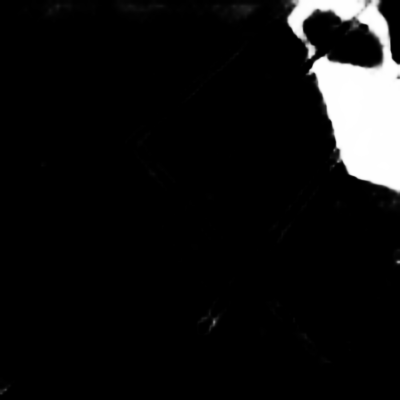}\\[0.1em]
      \includegraphics[width=0.15\textwidth, height=0.15\textwidth]{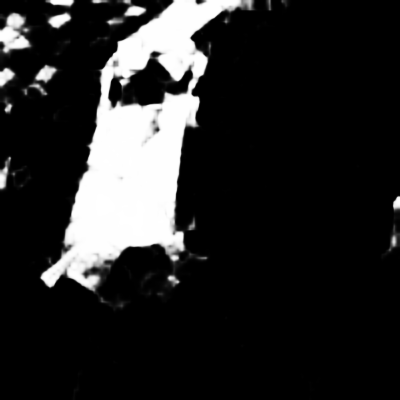}\\[0.1em]
      \includegraphics[width=0.15\textwidth, height=0.15\textwidth]{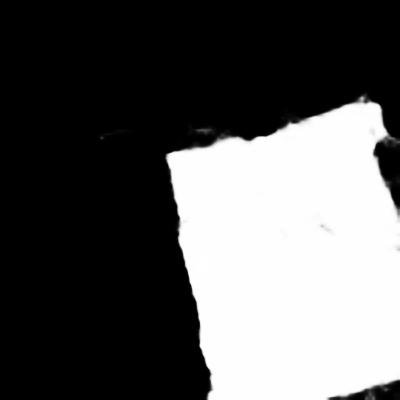}
    \end{tabular}}%
  \hspace{0.1em}%
  \subcaptionbox{\scriptsize{DSD}}{%
    \begin{tabular}{c}
      \includegraphics[width=0.15\textwidth, height=0.15\textwidth]{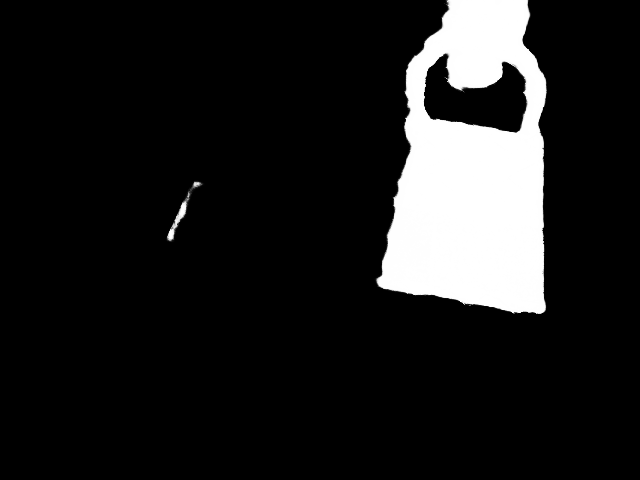} \\[0.1em]
      \includegraphics[width=0.15\textwidth, height=0.15\textwidth]{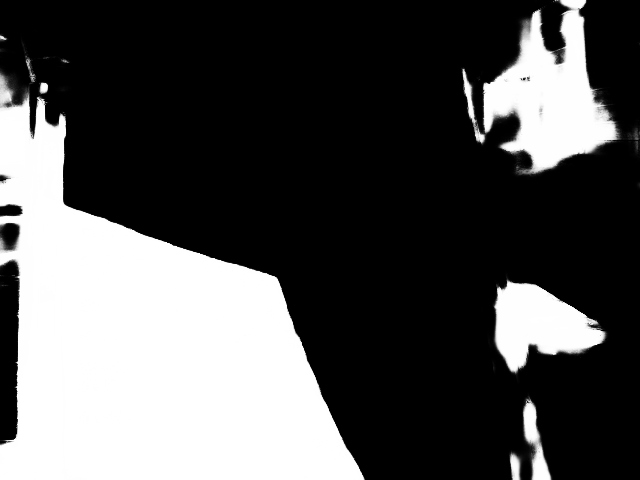} \\[0.1em]
      \includegraphics[width=0.15\textwidth, height=0.15\textwidth]{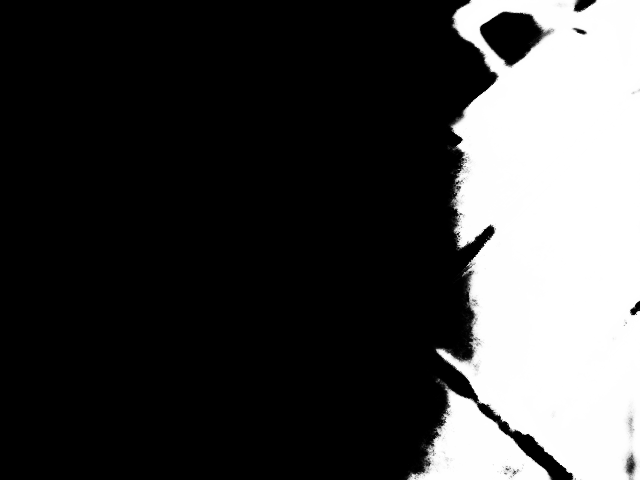}\\[0.1em]
      \includegraphics[width=0.15\textwidth, height=0.15\textwidth]{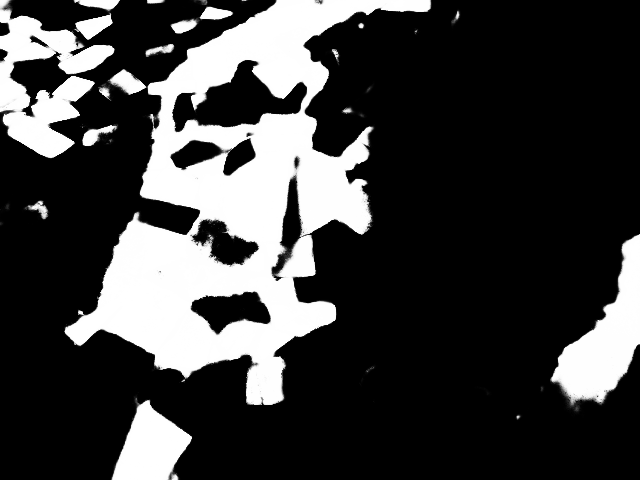}\\[0.1em]
      \includegraphics[width=0.15\textwidth, height=0.15\textwidth]{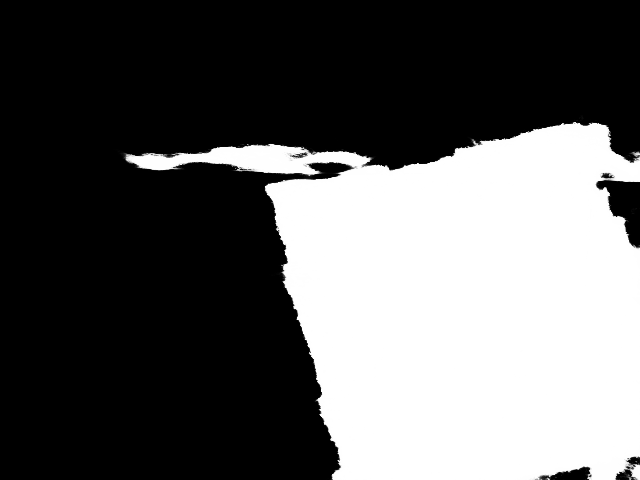}
    \end{tabular}}%
  \hspace{0.1em}%
  \subcaptionbox{\scriptsize{MTMT}}{%
    \begin{tabular}{c}
      \includegraphics[width=0.15\textwidth, height=0.15\textwidth]{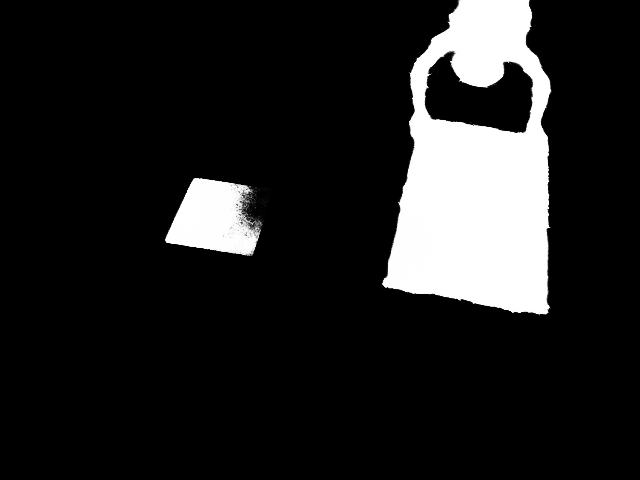} \\[0.1em]
      \includegraphics[width=0.15\textwidth, height=0.15\textwidth]{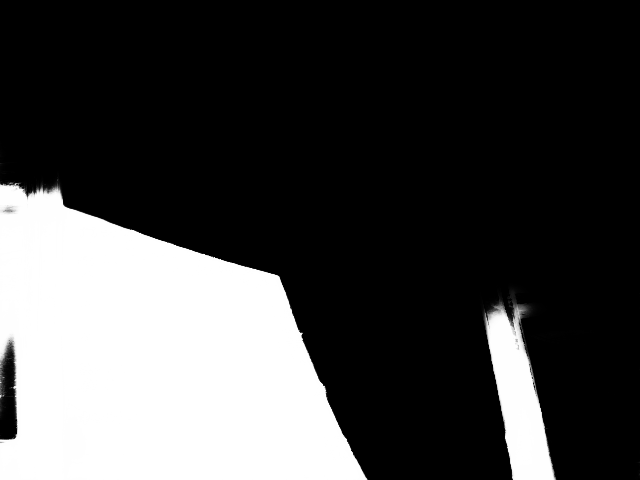} \\[0.1em]
      \includegraphics[width=0.15\textwidth, height=0.15\textwidth]{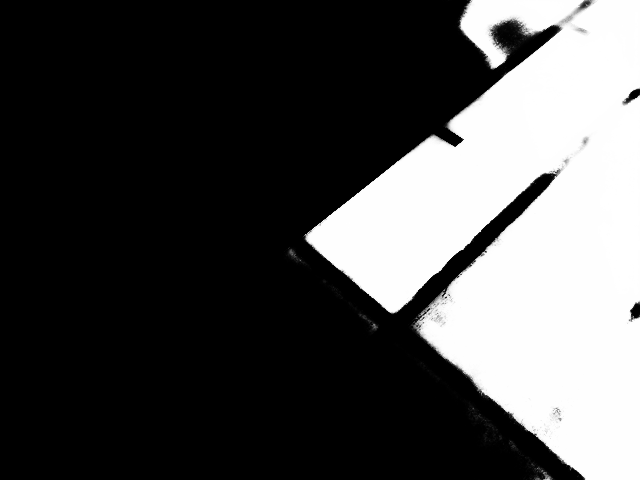}\\[0.1em]
      \includegraphics[width=0.15\textwidth, height=0.15\textwidth]{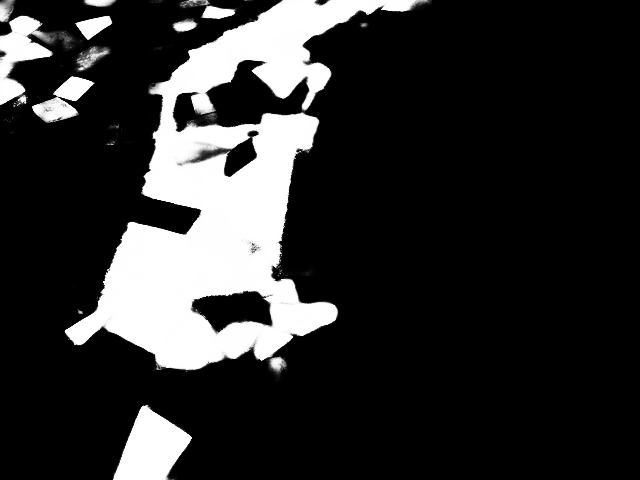}\\[0.1em]
      \includegraphics[width=0.15\textwidth, height=0.15\textwidth]{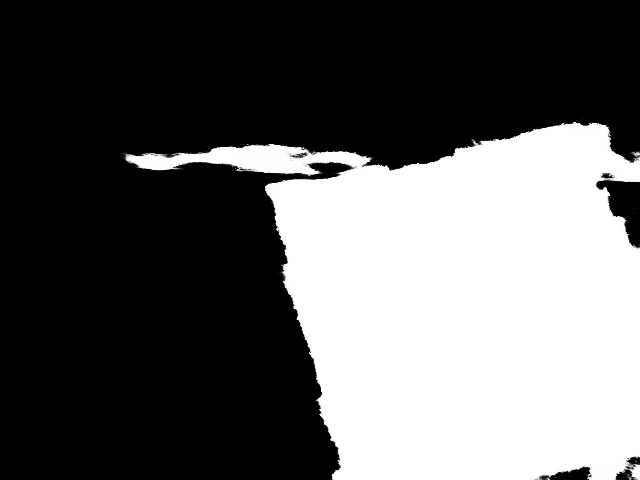}
    \end{tabular}}%
  \hspace{0.1em}%
  \subcaptionbox{\scriptsize{FDRNet}}{%
    \begin{tabular}{c}
      \includegraphics[width=0.15\textwidth, height=0.15\textwidth]{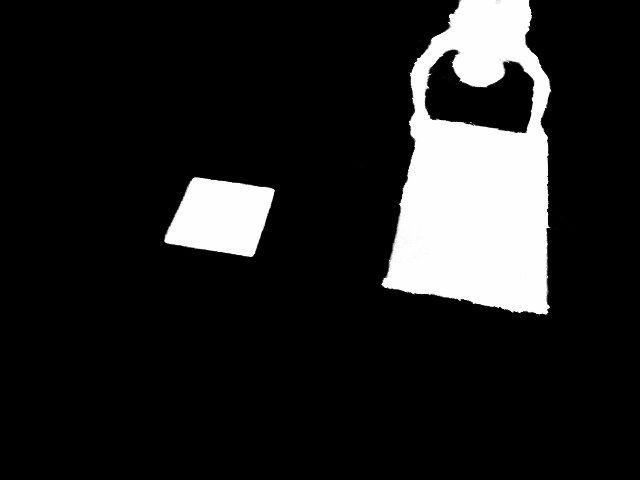} \\[0.1em]
      \includegraphics[width=0.15\textwidth, height=0.15\textwidth]{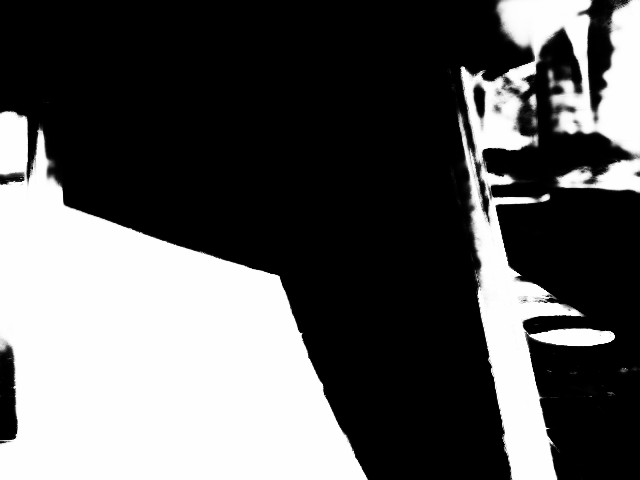} \\[0.1em]
      \includegraphics[width=0.15\textwidth, height=0.15\textwidth]{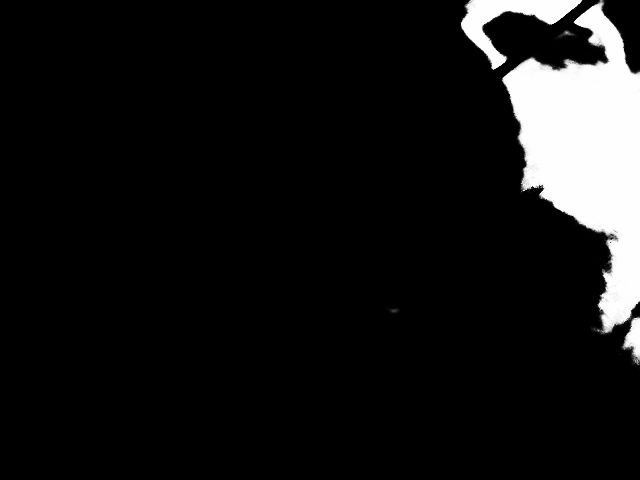}\\[0.1em]
      \includegraphics[width=0.15\textwidth, height=0.15\textwidth]{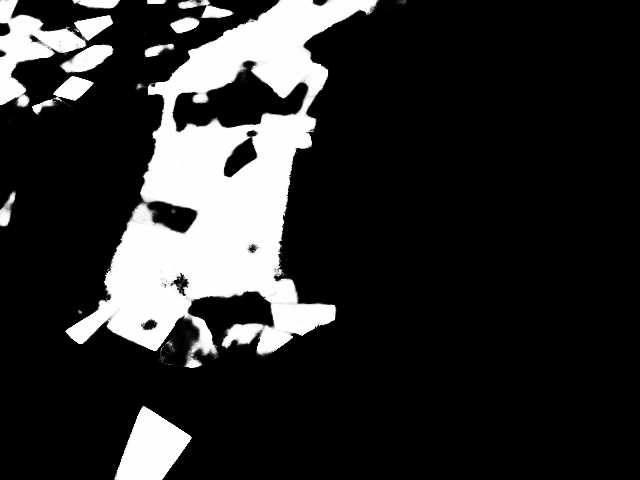}\\[0.1em]
      \includegraphics[width=0.15\textwidth, height=0.15\textwidth]{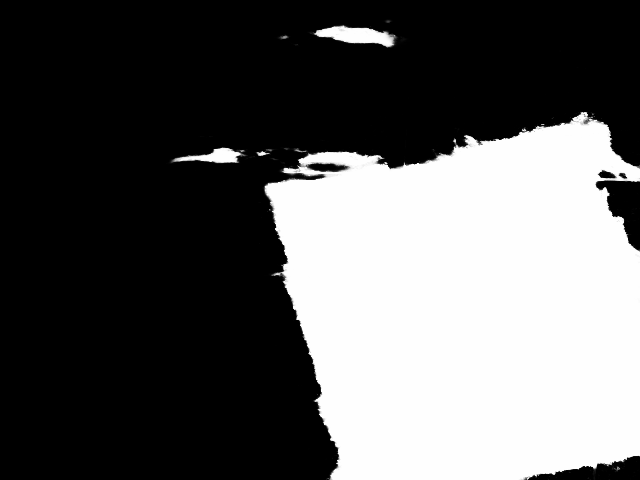}
    \end{tabular}}%
  \hspace{0.0em}%
  
\caption{More results on ISTD~\cite{wang2018stacked} dataset for shadow detection. We compare to DSD~\cite{zhao2021self}, MTMT~\cite{mtmt}, FDRNet~\cite{zhu2021mitigating}.}
\label{fig:supp_shadow}
\end{figure*}

\begin{figure*}[t]
  \captionsetup[subfigure]{position=b}
  \centering
  \setlength{\tabcolsep}{0pt}
  \subcaptionbox{\scriptsize{Input}}{%
    \begin{tabular}{c}
      \includegraphics[width=0.15\textwidth, height=0.15\textwidth]{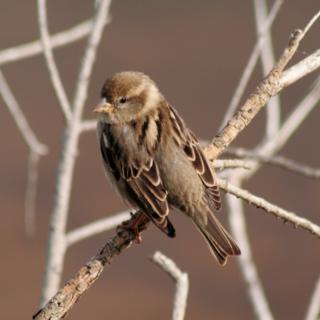} \\[0.1em]
      \includegraphics[width=0.15\textwidth, height=0.15\textwidth]{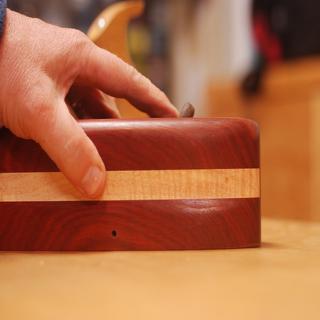} \\[0.1em]
      \includegraphics[width=0.15\textwidth, height=0.15\textwidth]{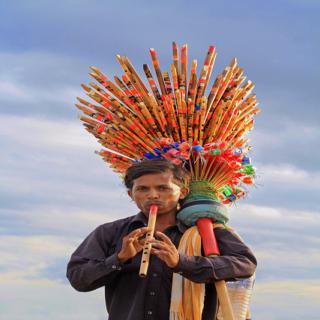} \\[0.1em]
      \includegraphics[width=0.15\textwidth, height=0.15\textwidth]{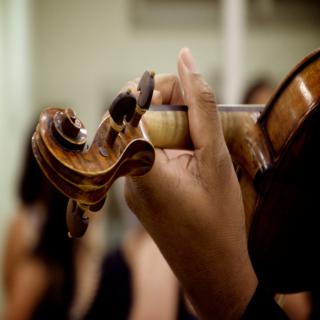} \\[0.1em]
      \includegraphics[width=0.15\textwidth, height=0.15\textwidth]{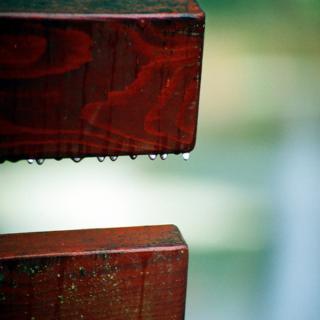}
    \end{tabular}}%
  \hspace{0.1em}%
  \subcaptionbox{\scriptsize{GT}}{%
    \begin{tabular}{c}
      \includegraphics[width=0.15\textwidth, height=0.15\textwidth]{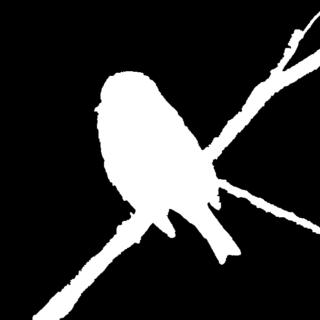} \\[0.1em]
      \includegraphics[width=0.15\textwidth, height=0.15\textwidth]{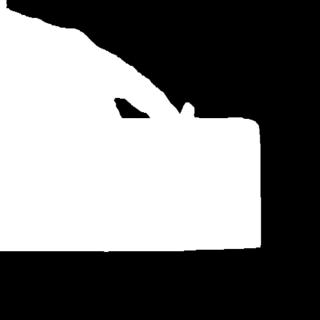} \\[0.1em]
      \includegraphics[width=0.15\textwidth, height=0.15\textwidth]{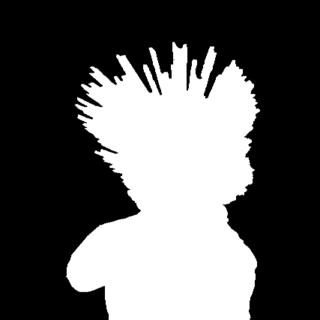} \\[0.1em]
      \includegraphics[width=0.15\textwidth, height=0.15\textwidth]{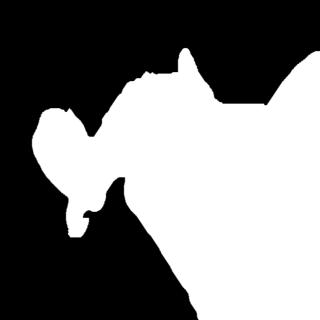} \\[0.1em]
      \includegraphics[width=0.15\textwidth, height=0.15\textwidth]{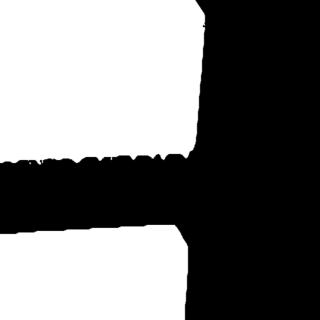}
    \end{tabular}}%
  \hspace{0.1em}%
  \subcaptionbox{\scriptsize{Ours}}{%
    \begin{tabular}{c}
      \includegraphics[width=0.15\textwidth, height=0.15\textwidth]{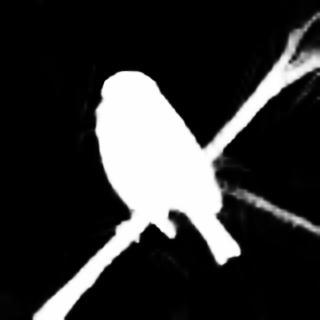} \\[0.1em]
      \includegraphics[width=0.15\textwidth, height=0.15\textwidth]{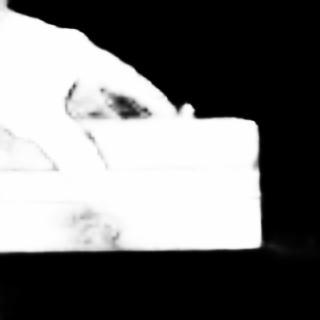} \\[0.1em]
      \includegraphics[width=0.15\textwidth, height=0.15\textwidth]{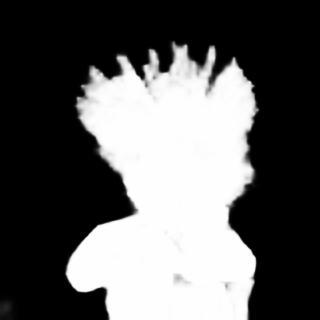} \\[0.1em]
      \includegraphics[width=0.15\textwidth, height=0.15\textwidth]{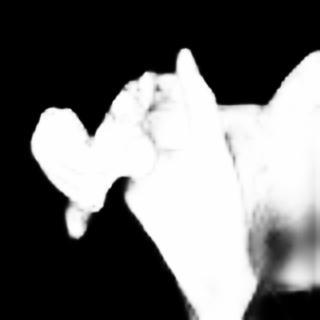} \\[0.1em]
      \includegraphics[width=0.15\textwidth, height=0.15\textwidth]{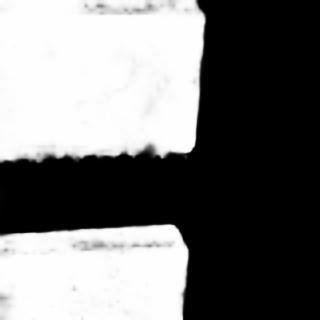}
    \end{tabular}}%
  \hspace{0.1em}%
  \subcaptionbox{\scriptsize{BTBNet}}{%
    \begin{tabular}{c}
      \includegraphics[width=0.15\textwidth, height=0.15\textwidth]{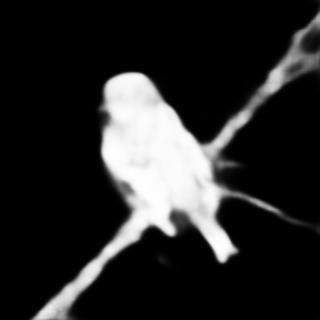} \\[0.1em]
      \includegraphics[width=0.15\textwidth, height=0.15\textwidth]{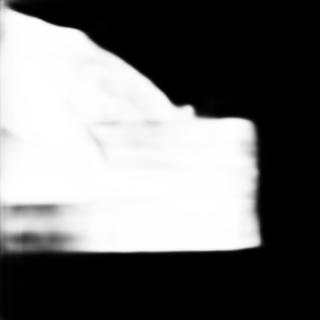} \\[0.1em]
      \includegraphics[width=0.15\textwidth, height=0.15\textwidth]{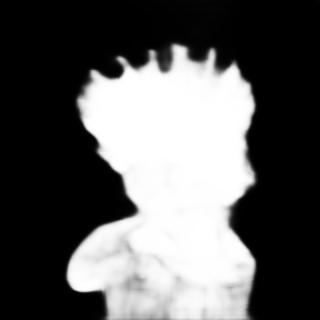} \\[0.1em]
      \includegraphics[width=0.15\textwidth, height=0.15\textwidth]{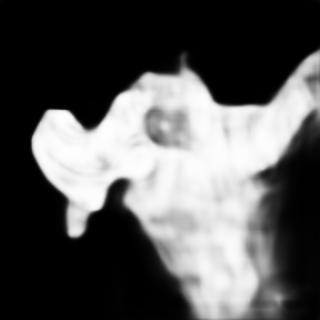} \\[0.1em]
      \includegraphics[width=0.15\textwidth, height=0.15\textwidth]{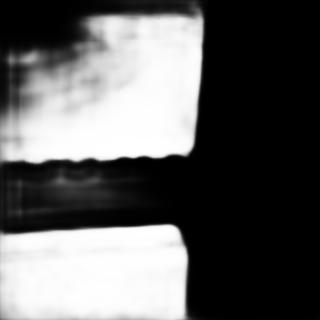}
    \end{tabular}}%
  \hspace{0.1em}%
  \subcaptionbox{\scriptsize{CENet}}{%
    \begin{tabular}{c}
      \includegraphics[width=0.15\textwidth, height=0.15\textwidth]{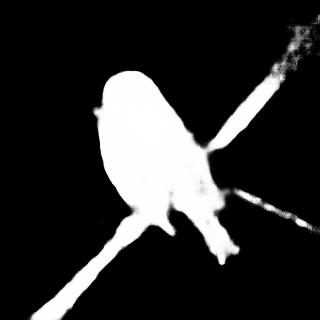} \\[0.1em]
      \includegraphics[width=0.15\textwidth, height=0.15\textwidth]{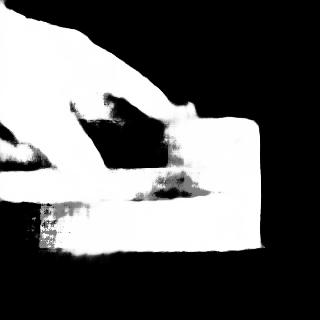} \\[0.1em]
      \includegraphics[width=0.15\textwidth, height=0.15\textwidth]{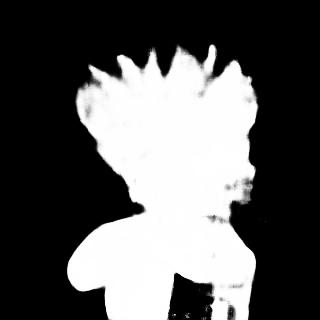} \\[0.1em]
      \includegraphics[width=0.15\textwidth, height=0.15\textwidth]{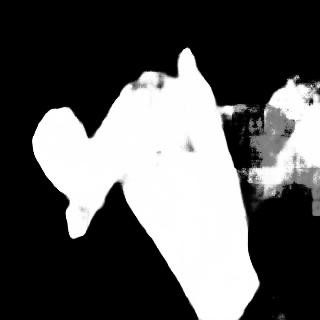} \\[0.1em]
      \includegraphics[width=0.15\textwidth, height=0.15\textwidth]{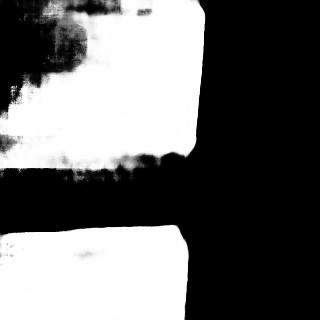}
    \end{tabular}}%
  \hspace{0.1em}%
  \subcaptionbox{\scriptsize{EFENet}}{%
    \begin{tabular}{c}
      \includegraphics[width=0.15\textwidth, height=0.15\textwidth]{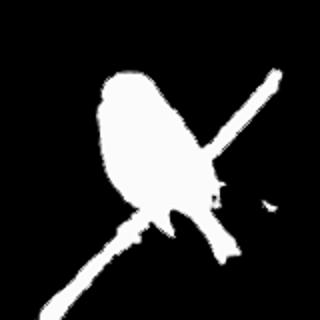} \\[0.1em]
      \includegraphics[width=0.15\textwidth, height=0.15\textwidth]{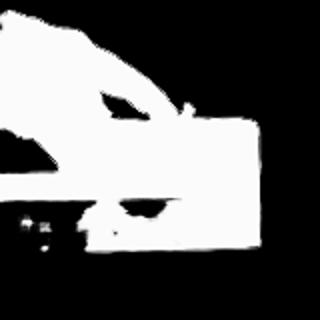} \\[0.1em]
      \includegraphics[width=0.15\textwidth, height=0.15\textwidth]{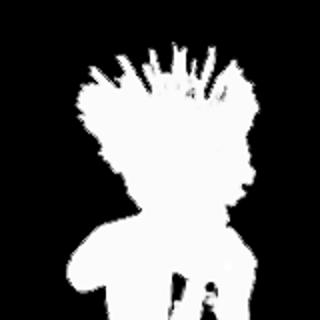} \\[0.1em]
      \includegraphics[width=0.15\textwidth, height=0.15\textwidth]{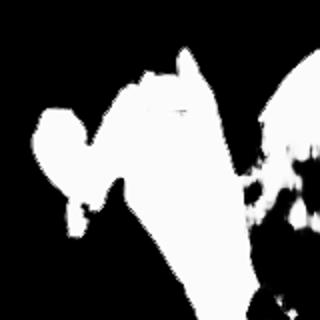} \\[0.1em]
      \includegraphics[width=0.15\textwidth, height=0.15\textwidth]{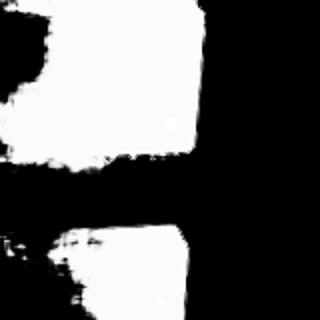}
    \end{tabular}}%
  \hspace{0.0em}%
  
\caption{More results on CUHK~\cite{shi2014discriminative} dataset for defocus blur detection.We compare to BTBNet~\cite{zhao2019btbnet}, CENet~\cite{zhao2019cenet} and EFENet~\cite{zhao2021defocus}.}
\label{fig:supp_defocus}
\end{figure*}

\begin{figure*}[t]
  \captionsetup[subfigure]{position=b}
  \centering
  \setlength{\tabcolsep}{0pt}
  \subcaptionbox{\scriptsize{Input}}{%
    \begin{tabular}{c}
      \includegraphics[width=0.13\textwidth, height=0.13\textwidth]{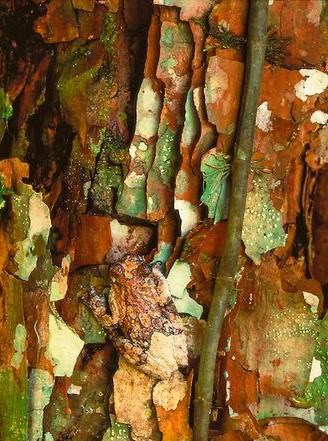} \\[0.1em]
      \includegraphics[width=0.13\textwidth, height=0.13\textwidth]{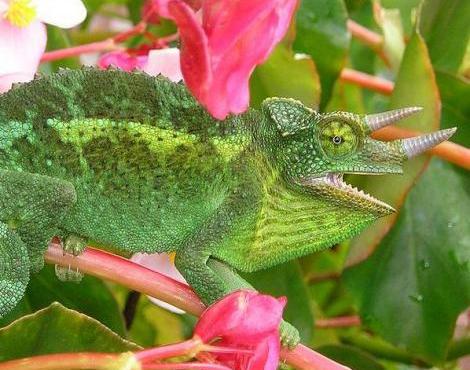} \\[0.1em]
      \includegraphics[width=0.13\textwidth, height=0.13\textwidth]{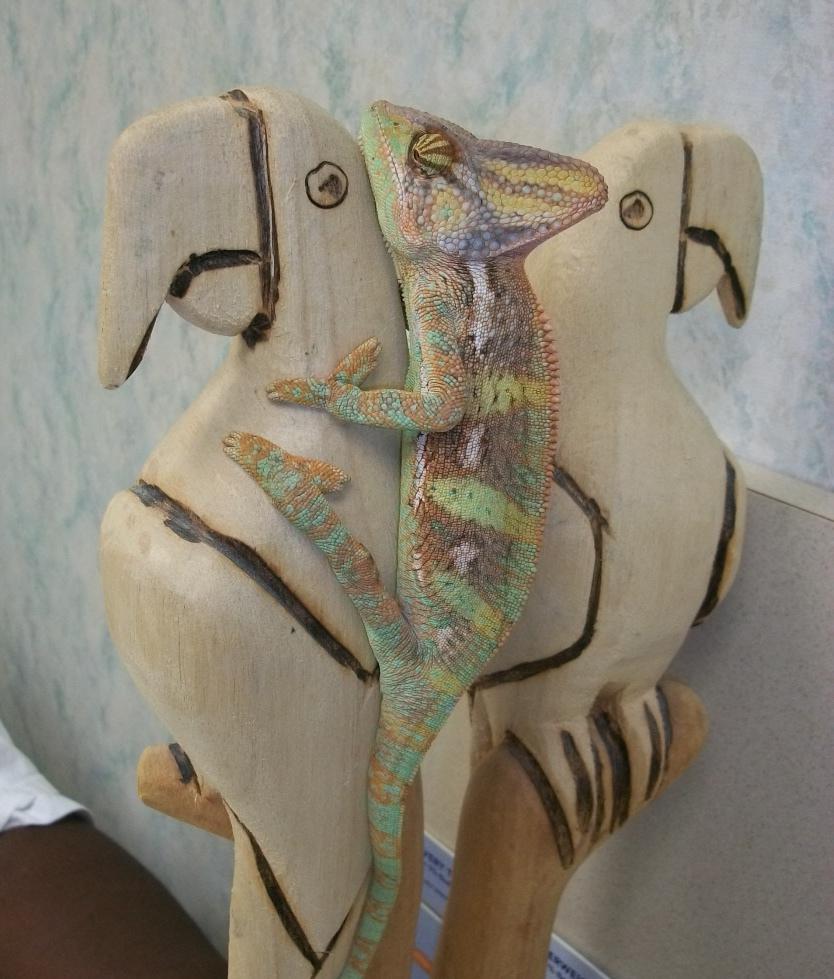}\\[0.1em]
      \includegraphics[width=0.13\textwidth, height=0.13\textwidth]{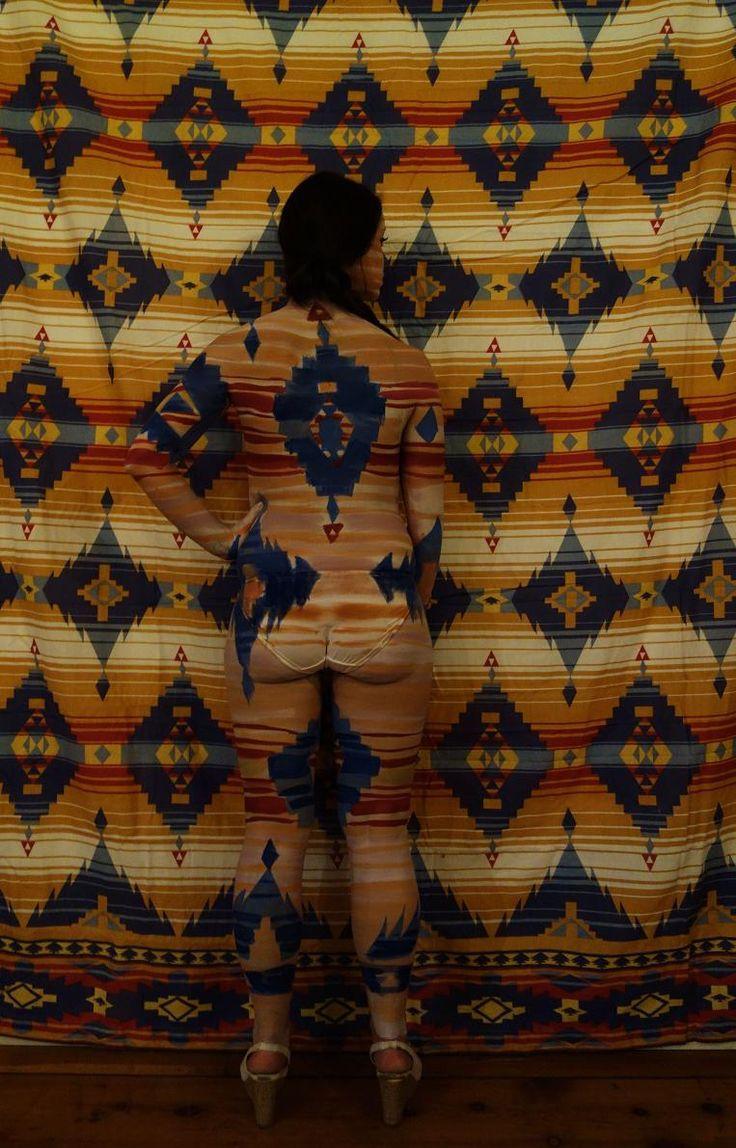}\\[0.1em]
      \includegraphics[width=0.13\textwidth, height=0.13\textwidth]{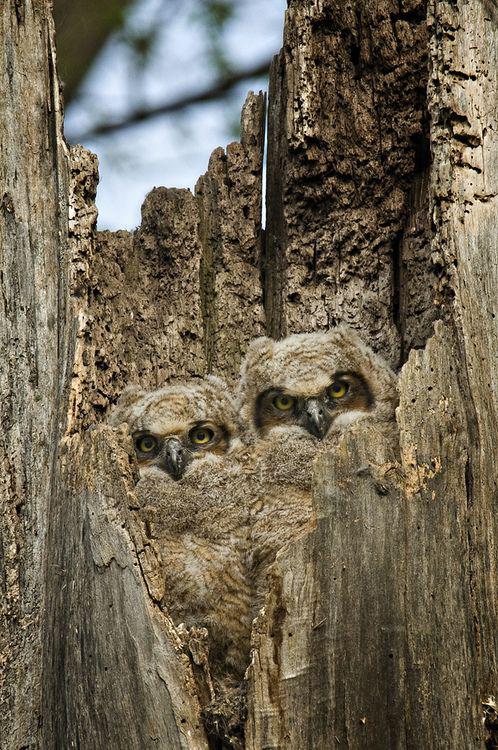}
    \end{tabular}}%
  \hspace{0.1em}%
  \subcaptionbox{\scriptsize{GT}}{%
    \begin{tabular}{c}
      \includegraphics[width=0.13\textwidth, height=0.13\textwidth]{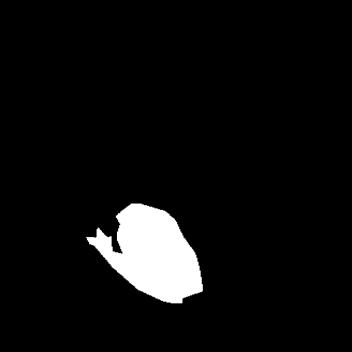} \\[0.1em]
      \includegraphics[width=0.13\textwidth, height=0.13\textwidth]{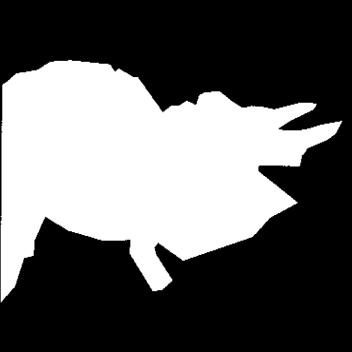} \\[0.1em]
      \includegraphics[width=0.13\textwidth, height=0.13\textwidth]{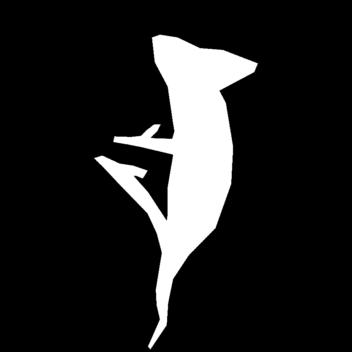}\\[0.1em]
      \includegraphics[width=0.13\textwidth, height=0.13\textwidth]{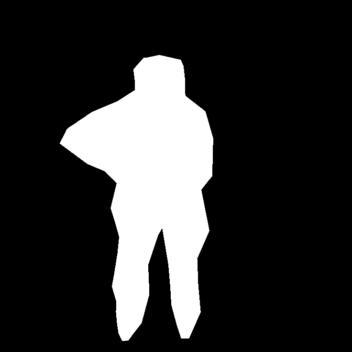}\\[0.1em]
      \includegraphics[width=0.13\textwidth, height=0.13\textwidth]{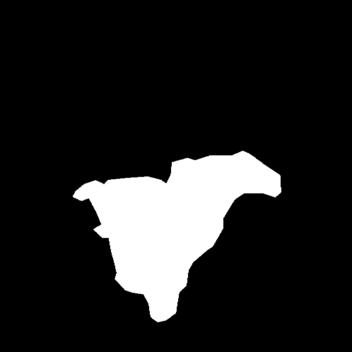}
    \end{tabular}}%
  \hspace{0.1em}%
  \subcaptionbox{\scriptsize{Ours}}{%
    \begin{tabular}{c}
      \includegraphics[width=0.13\textwidth, height=0.13\textwidth]{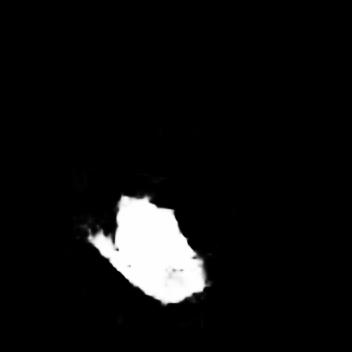} \\[0.1em]
      \includegraphics[width=0.13\textwidth, height=0.13\textwidth]{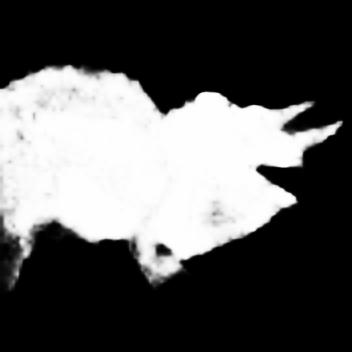} \\[0.1em]
      \includegraphics[width=0.13\textwidth, height=0.13\textwidth]{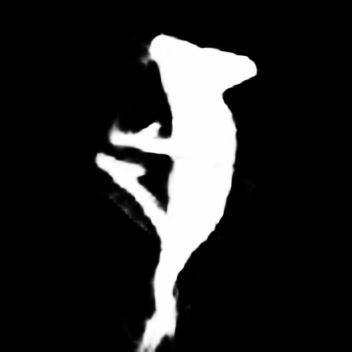}\\[0.1em]
      \includegraphics[width=0.13\textwidth, height=0.13\textwidth]{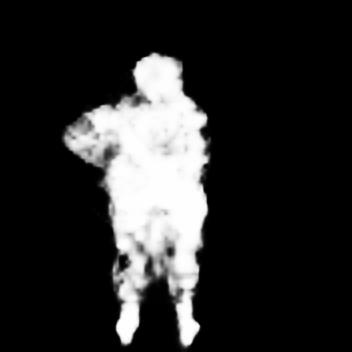}\\[0.1em]
      \includegraphics[width=0.13\textwidth, height=0.13\textwidth]{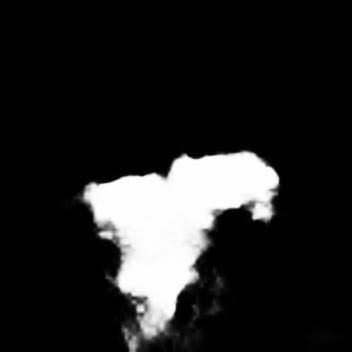}
    \end{tabular}}%
  \hspace{0.1em}%
  \subcaptionbox{\scriptsize{SINet}}{%
    \begin{tabular}{c}
      \includegraphics[width=0.13\textwidth, height=0.13\textwidth]{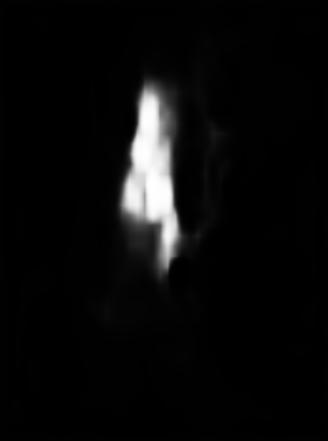} \\[0.1em]
      \includegraphics[width=0.13\textwidth, height=0.13\textwidth]{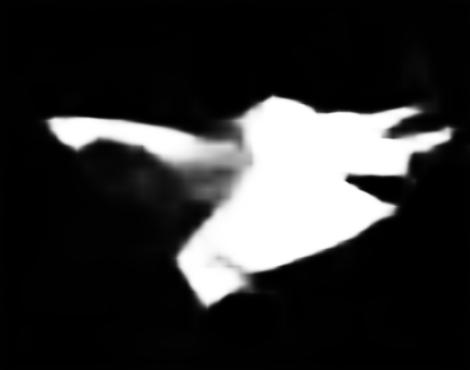} \\[0.1em]
      \includegraphics[width=0.13\textwidth, height=0.13\textwidth]{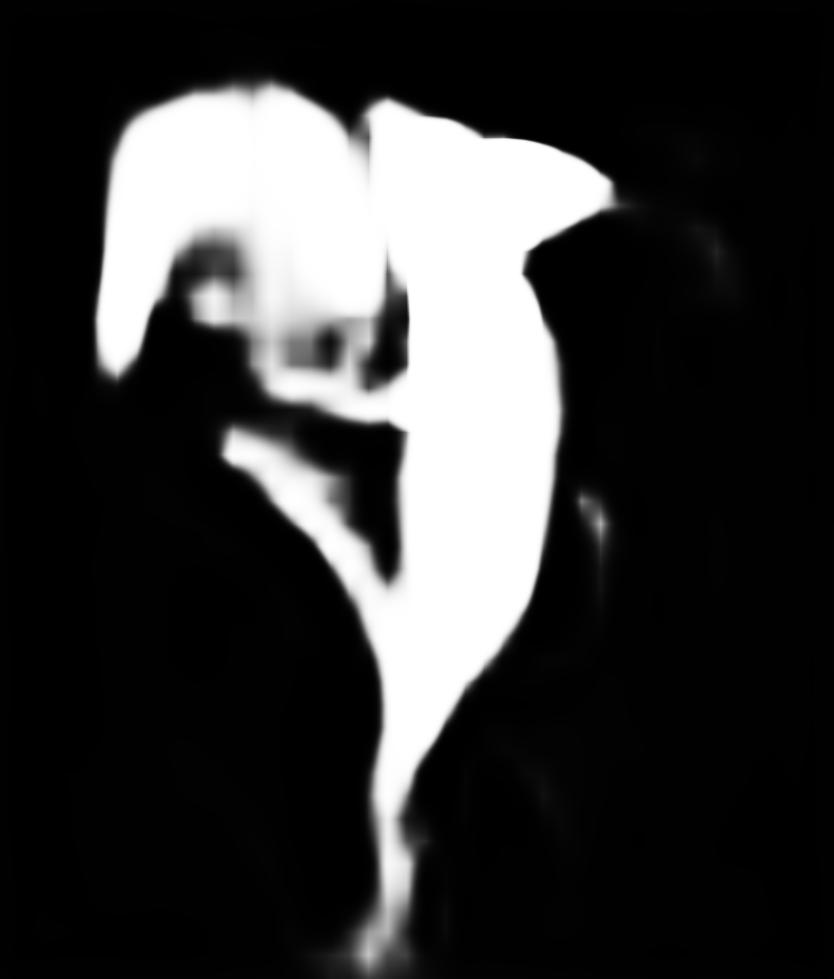}\\[0.1em]
      \includegraphics[width=0.13\textwidth, height=0.13\textwidth]{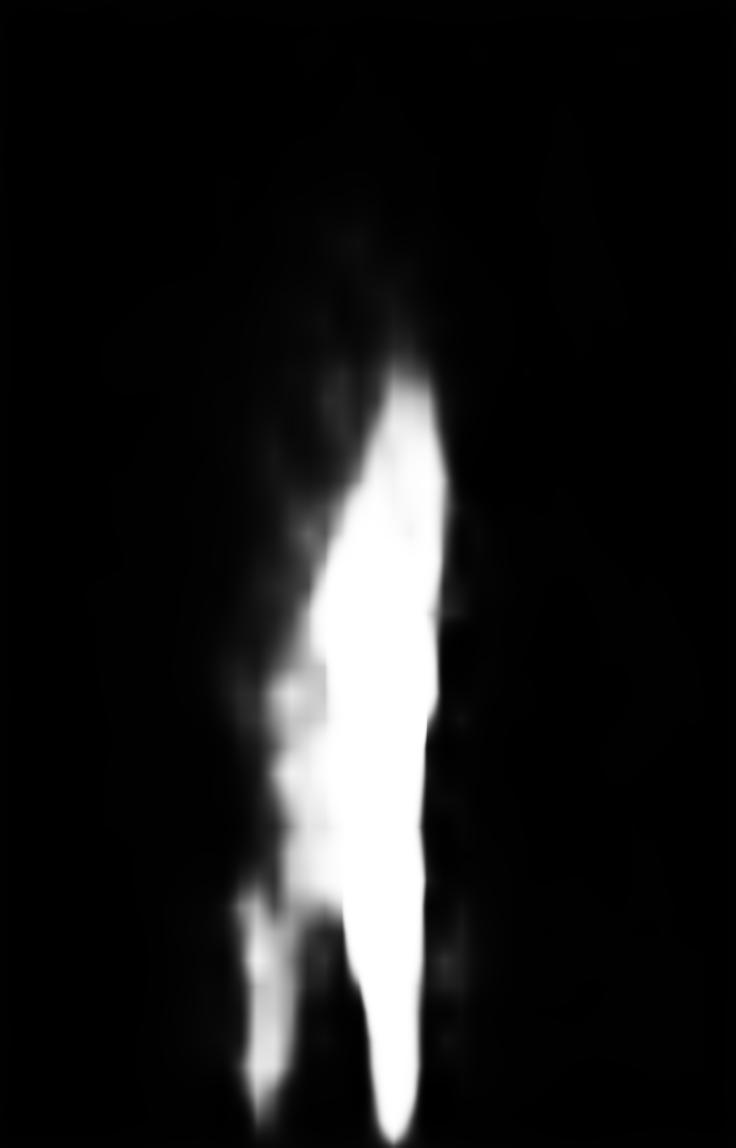}\\[0.1em]
      \includegraphics[width=0.13\textwidth, height=0.13\textwidth]{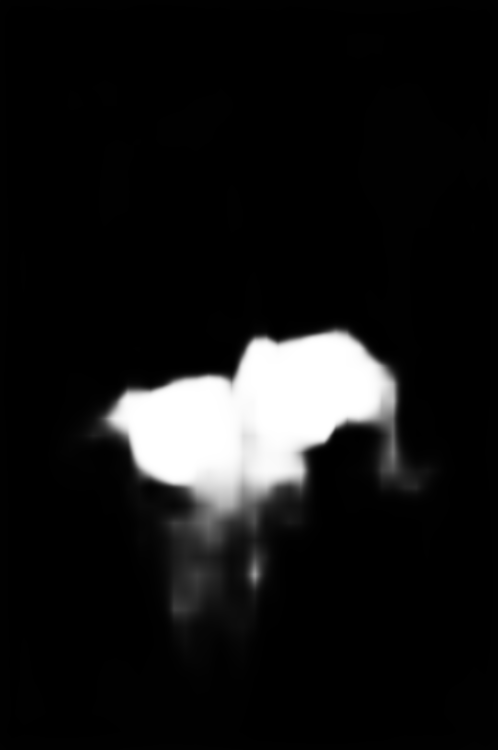}
    \end{tabular}}%
  \hspace{0.1em}%
  \subcaptionbox{\scriptsize{PFNet}}{%
    \begin{tabular}{c}
      \includegraphics[width=0.13\textwidth, height=0.13\textwidth]{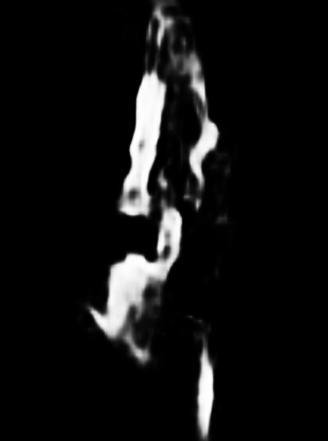} \\[0.1em]
      \includegraphics[width=0.13\textwidth, height=0.13\textwidth]{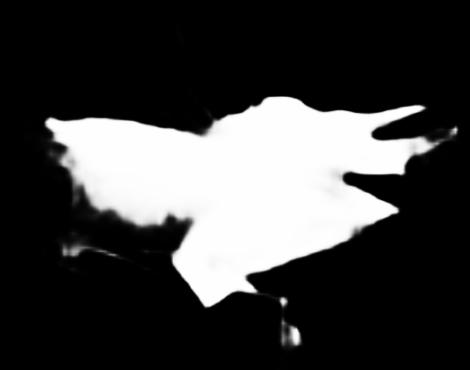} \\[0.1em]
      \includegraphics[width=0.13\textwidth, height=0.13\textwidth]{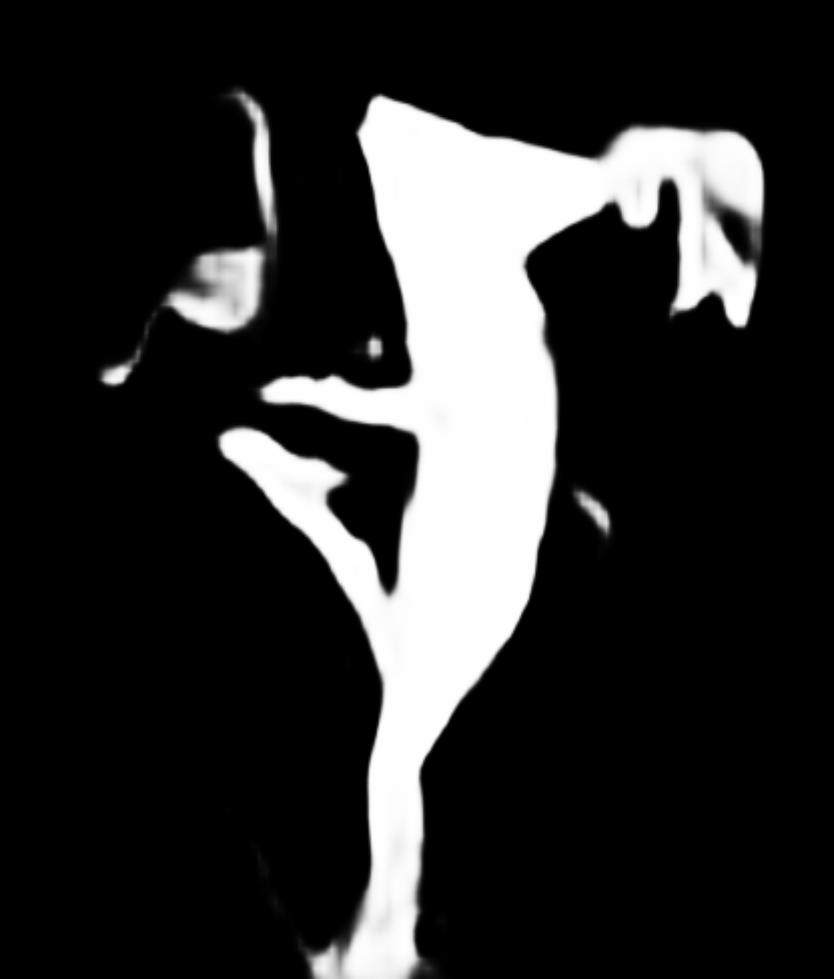}\\[0.1em]
      \includegraphics[width=0.13\textwidth, height=0.13\textwidth]{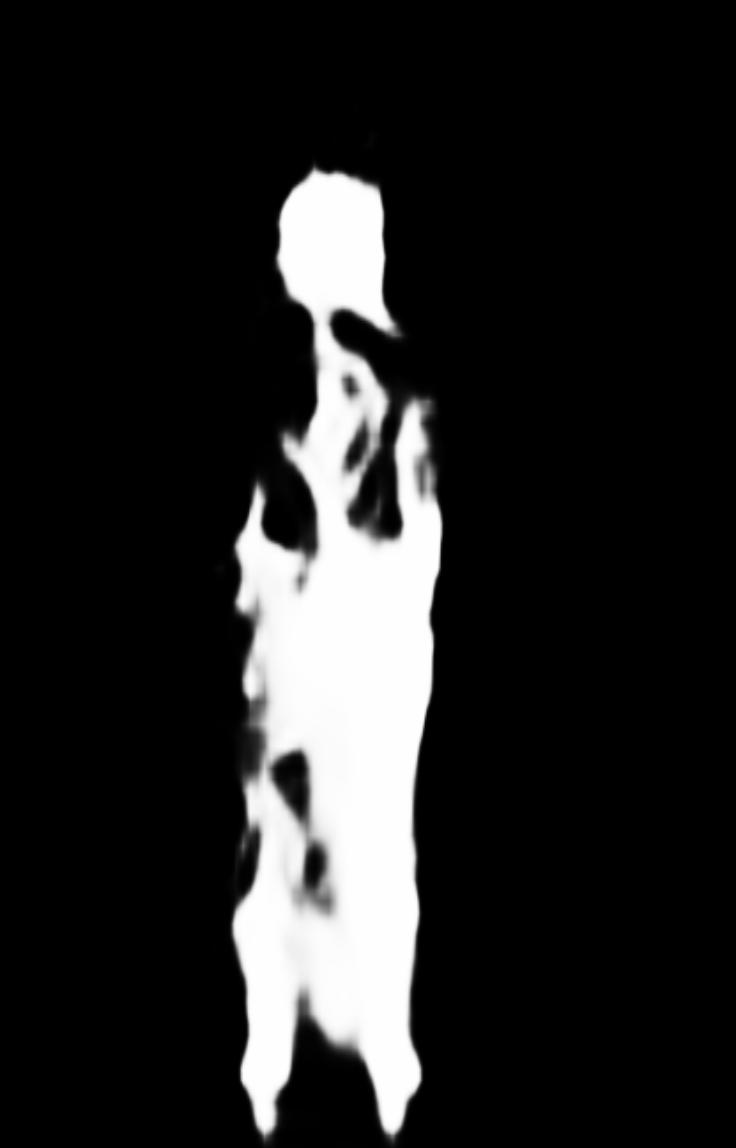}\\[0.1em]
      \includegraphics[width=0.13\textwidth, height=0.13\textwidth]{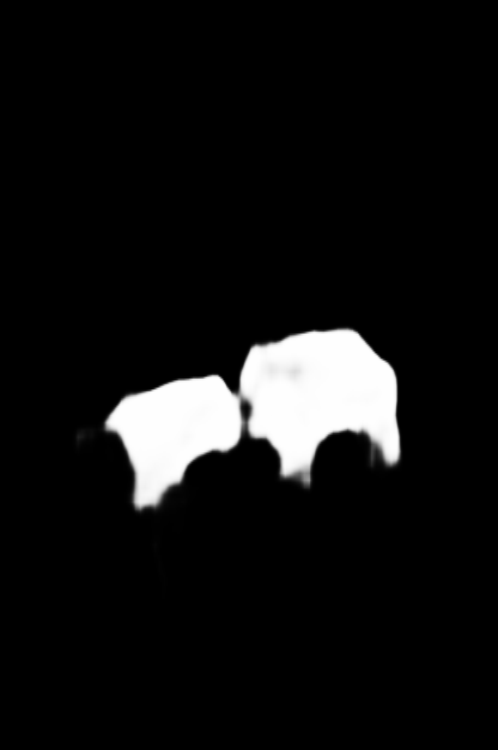}
    \end{tabular}}%
  \hspace{0.1em}%
  \subcaptionbox{\scriptsize{JCOD}}{%
    \begin{tabular}{c}
      \includegraphics[width=0.13\textwidth, height=0.13\textwidth]{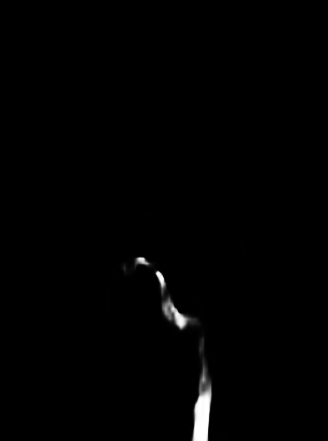} \\[0.1em]
      \includegraphics[width=0.13\textwidth, height=0.13\textwidth]{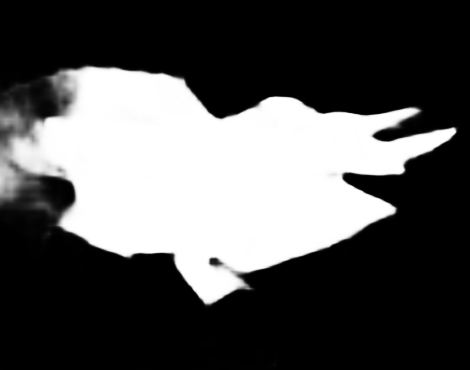} \\[0.1em]
      \includegraphics[width=0.13\textwidth, height=0.13\textwidth]{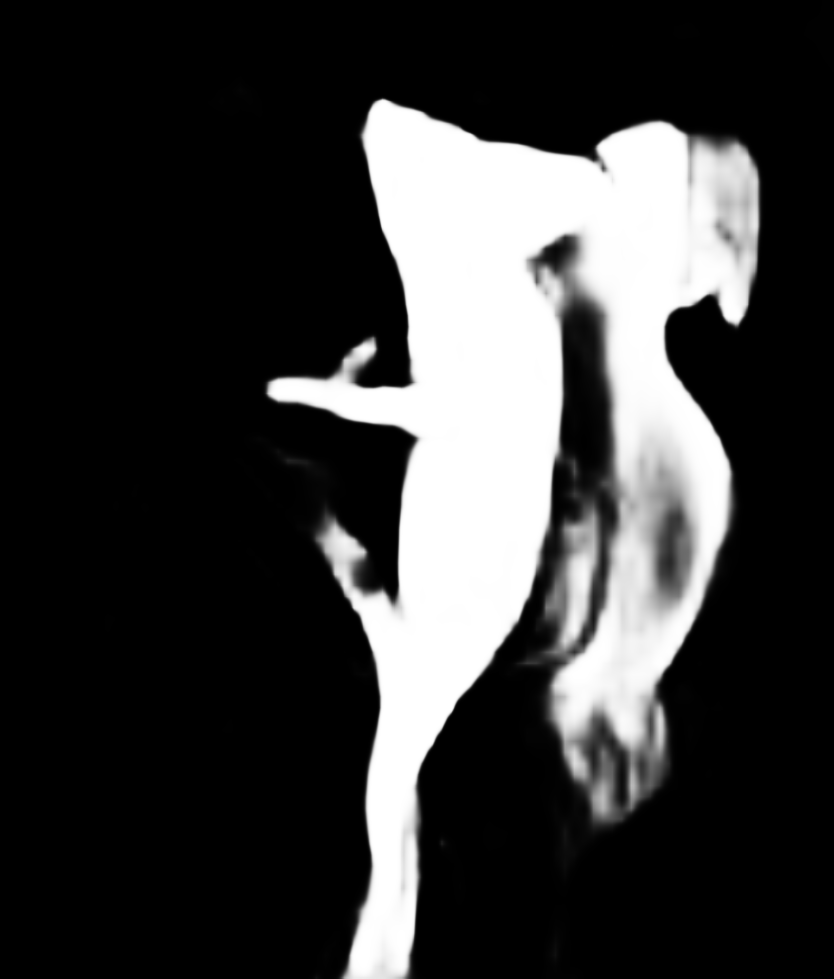}\\[0.1em]
      \includegraphics[width=0.13\textwidth, height=0.13\textwidth]{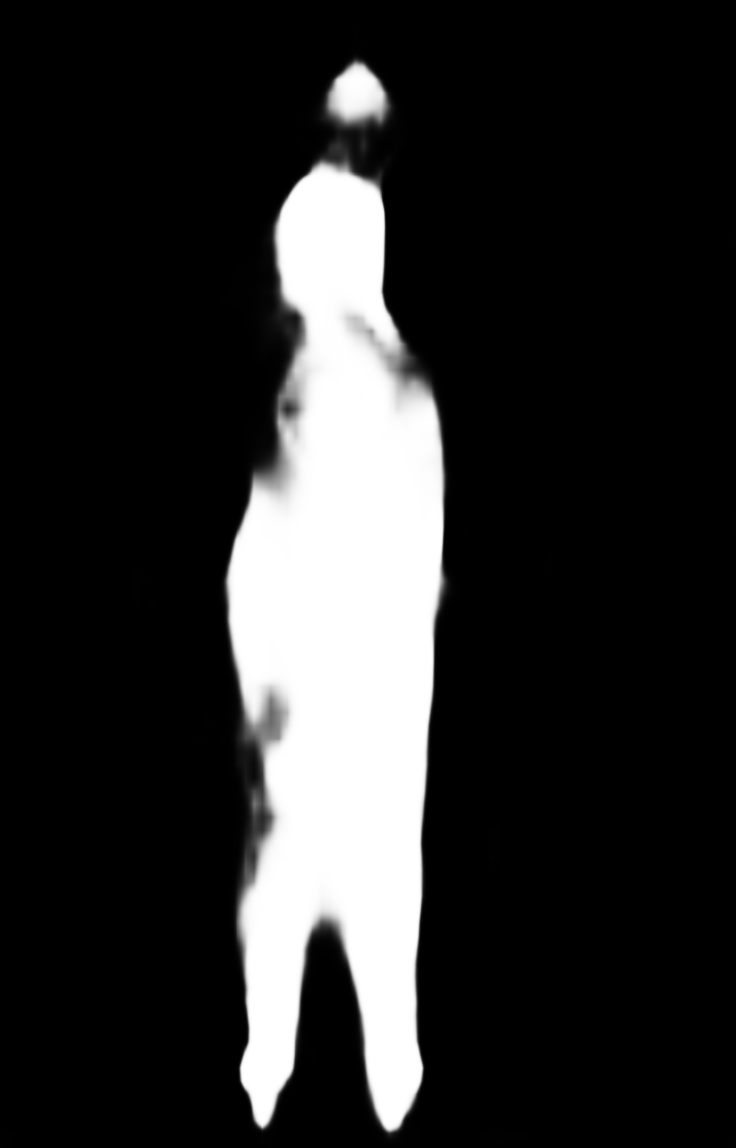}\\[0.1em]
      \includegraphics[width=0.13\textwidth, height=0.13\textwidth]{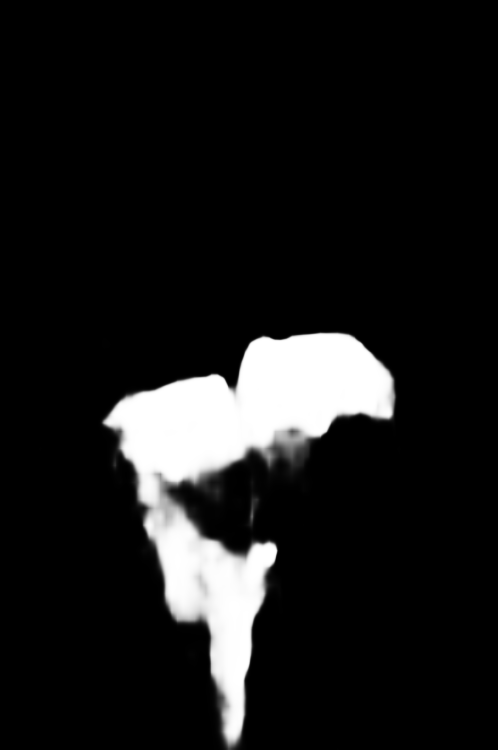}
    \end{tabular}}%
  \hspace{0.1em}%
  \subcaptionbox{\scriptsize{RankNet}}{%
    \begin{tabular}{c}
      \includegraphics[width=0.13\textwidth, height=0.13\textwidth]{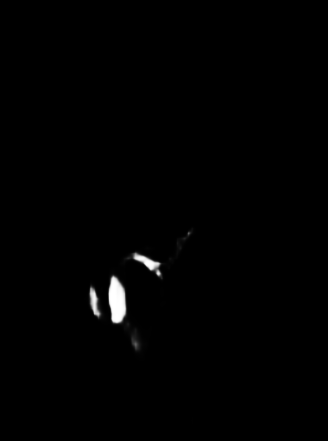} \\[0.1em]
      \includegraphics[width=0.13\textwidth, height=0.13\textwidth]{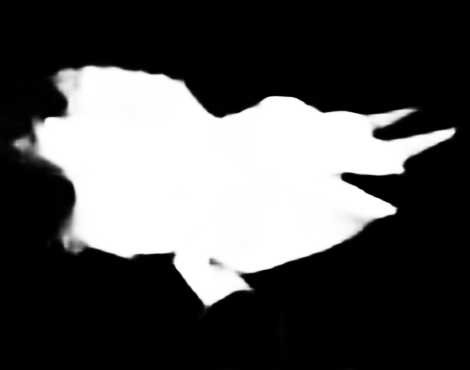} \\[0.1em]
      \includegraphics[width=0.13\textwidth, height=0.13\textwidth]{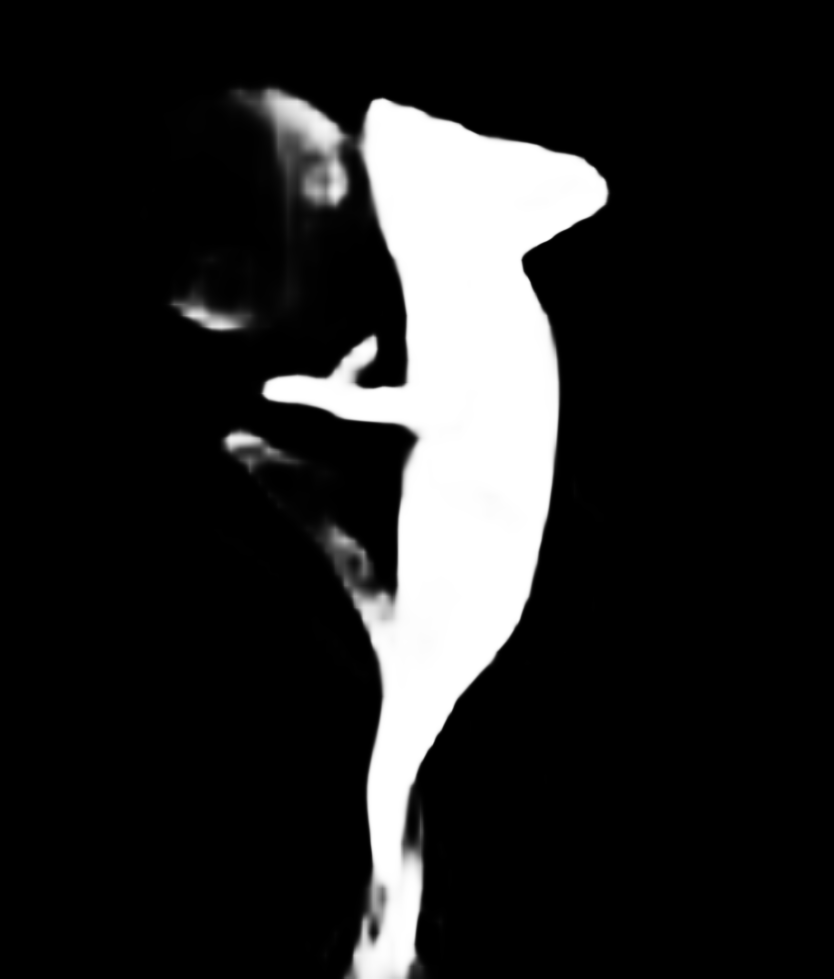}\\[0.1em]
      \includegraphics[width=0.13\textwidth, height=0.13\textwidth]{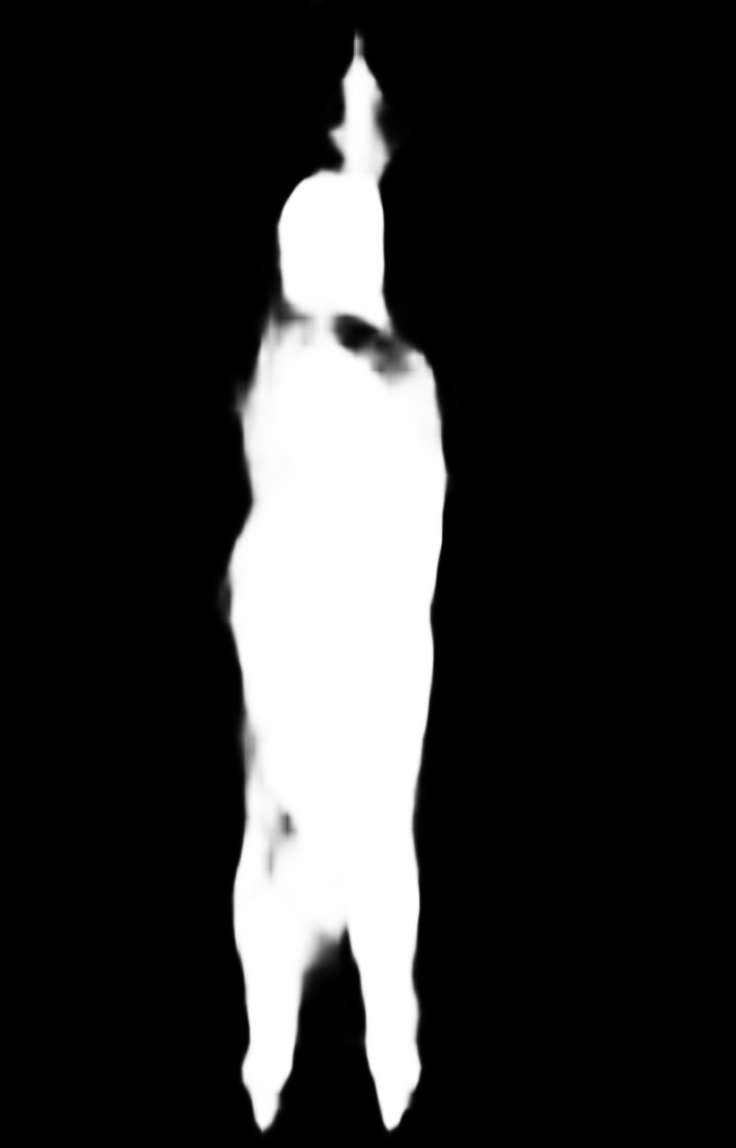}\\[0.1em]
      \includegraphics[width=0.13\textwidth, height=0.13\textwidth]{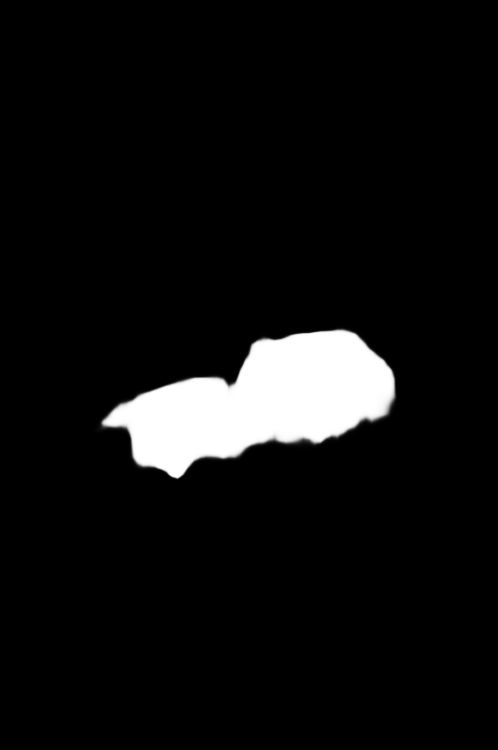}
    \end{tabular}}%
  \hspace{0.0em}%
  
\caption{More results on CAMO~\cite{le2019anabranch} dataset for camouflaged object detection. We compare to SINet~\cite{fan2020camouflaged}, PFNEt~\cite{mei2021camouflaged}, JCOD~\cite{li2021uncertainty} and RankNet~\cite{lv2021simultaneously}.}
\label{fig:supp_cod}
\end{figure*}
 \fi

\end{document}